\documentclass{article} 
\usepackage{iclr2026_conference}
\usepackage{times}
\iclrfinalcopy

\usepackage[table]{xcolor}
\usepackage[utf8]{inputenc} 
\usepackage[T1]{fontenc}    
\usepackage{hyperref}       
\usepackage{url}            
\usepackage{booktabs}       
\usepackage{amsfonts}       
\usepackage{nicefrac}       
\usepackage{microtype}      
\usepackage{xcolor}         
\usepackage{caption}      
\usepackage{hyperref}     
\usepackage{lipsum}       
\usepackage{adjustbox}
\usepackage{graphicx}
\usepackage{float}
\usepackage{xspace}
\usepackage{amsmath} 
\usepackage{cleveref}
\usepackage{natbib}
\usepackage{wrapfig}
\usepackage{makecell}
\usepackage{multirow}
\usepackage{enumitem}  
\usepackage{diagbox}
\definecolor{myblue}{RGB}{230, 240, 255}

\title{Understanding vs. Generation: Navigating Optimization Dilemma in Multimodal Models}

\author{
Sen Ye$^{1, 2}$ \quad
Mengde Xu$^{2}$ \quad
Shuyang Gu$^{2} \footnotemark[2]$ \quad
Di He$^{1}$ \quad
Liwei Wang$^{1,3,4} \footnotemark[2]$ \quad
Han Hu$^{2}$ \quad
\\[0.5em]
{\small \textsuperscript{1} State Key Laboratory of General Artificial Intelligence, Peking University}  \ \ \
{\small \textsuperscript{2} Tencent}\\
{\small \textsuperscript{3} Center for Data Science, Peking University}
{\small \textsuperscript{4} Center for Machine Learning Research, Peking University}
}

\begin{document}

\maketitle
\footnotetext[2]{Corresponding Author.}

\begin{abstract}
Current research in multimodal models faces a key challenge where enhancing generative capabilities often comes at the expense of understanding, and vice versa. We analyzed this trade-off and identify the primary cause might be the potential conflict between generation and understanding, which creates a competitive dynamic within the model. To address this, we propose the Reason-Reflect-Refine (R3) framework. This innovative algorithm re-frames the single-step generation task into a multi-step process of "generate-understand-regenerate". By explicitly leveraging the model's understanding capability during generation, we successfully mitigate the optimization dilemma, achieved stronger generation results and improved understanding ability which are related to the generation process. This offers valuable insights for designing next-generation unified multimodal models. Code is available at https://github.com/sen-ye/R3.
\end{abstract}

\section{Introduction}
\label{intro}
For decades, the pursuit of Artificial General Intelligence (AGI) has been a central goal in AI research. A key step toward this vision is the development of unified multimodal models capable of both understanding and generating visual information, much like humans do~\citep{team2023gemini,gpt4o}. However, recent advance in large-scale multimodal learning points to a core dilemma that understanding and generation are hard to improve simultaneously~\citep{henighan2020scaling,emu3,janus,zhang2025resolving}. For instance, models fine-tuned for high-fidelity image synthesis, such as those based on diffusion architectures, often struggle with tasks requiring precise visual understanding, including object counting or spatial reasoning~\citep{cho2023dall,geneval,tiif,an2026genius}. Conversely, models optimized for tasks like visual question answering (VQA) or dense captioning tend to exhibit weaker creative and generative performance compared to their generative counterparts~\citep{dong2023dreamllm,showo,janus}.

A number of efforts have been made to resolve this tension. Some researchers argue that different tasks require specialized tokenizers~\citep{janus,januspro}, and thus propose unified tokenization schemes~\citep{team2024chameleon,unitok} to harmonize representation across modalities. Others attempt to disentangle understanding and generation by designing novel architectures that allocate separate capacity to each function~\citep{liang2024mixture,bagel,zhang2025resolving}. While these approaches achieve partial success, we argue that the crux of the conflict comes from the different training objectives. The generative objective typically maximizes the likelihood of samples under the data distribution, a goal that can be optimized without the understanding ability. As a result, the model’s capacity may be monopolized, in competition with the robust understanding required. This raises a fundamental question: should generation actively incorporate the model’s understanding of the underlying semantics?

In this paper, we introduce the Reason-Reflect-Refine (R3) framework, which re-conceptualizes generation as a multi-step process rather than a single-shot mapping. Instead of treating generation and understanding as competing objectives, R3 explicitly integrates understanding into the generative loop. By first reasoning the user request and producing an initial draft, then reflecting whether this generated result meets the user request, and finally refining the output according to the reflection, the model transforms understanding from a passive evaluation task into an active component of generation.

Analogous to a painter's creative process, our framework unfolds in distinct stages. In the Reason stage, the model first analyzes the user's intent to conceptualize the final image. It enriches the initial prompt by imagining and incorporating various fine-grained details, producing an explicit textual blueprint before synthesizing an initial draft.
Recognizing that high-fidelity generation from complex prompts is rarely achievable in a single attempt, the framework then enters an iterative Reflect–Refine loop. Here, the model evaluates its output against the original prompt—a process that demands strong multimodal understanding. If the output aligns well, the procedure terminates; otherwise, the model formulates corrective textual instructions and refines the image accordingly. This self-correction cycle continues until satisfactory alignment is reached, with the model itself deciding when to stop. The entire process is trained end-to-end using an outcome-based reward signal derived from the final image quality. To further enhance efficiency, we introduce an immediate rollout strategy that accelerates convergence without compromising performance.

\begin{wrapfigure}[18]{r}{0.5\textwidth}
    \vspace{-1em}
    \centering
    \includegraphics[width=\linewidth]{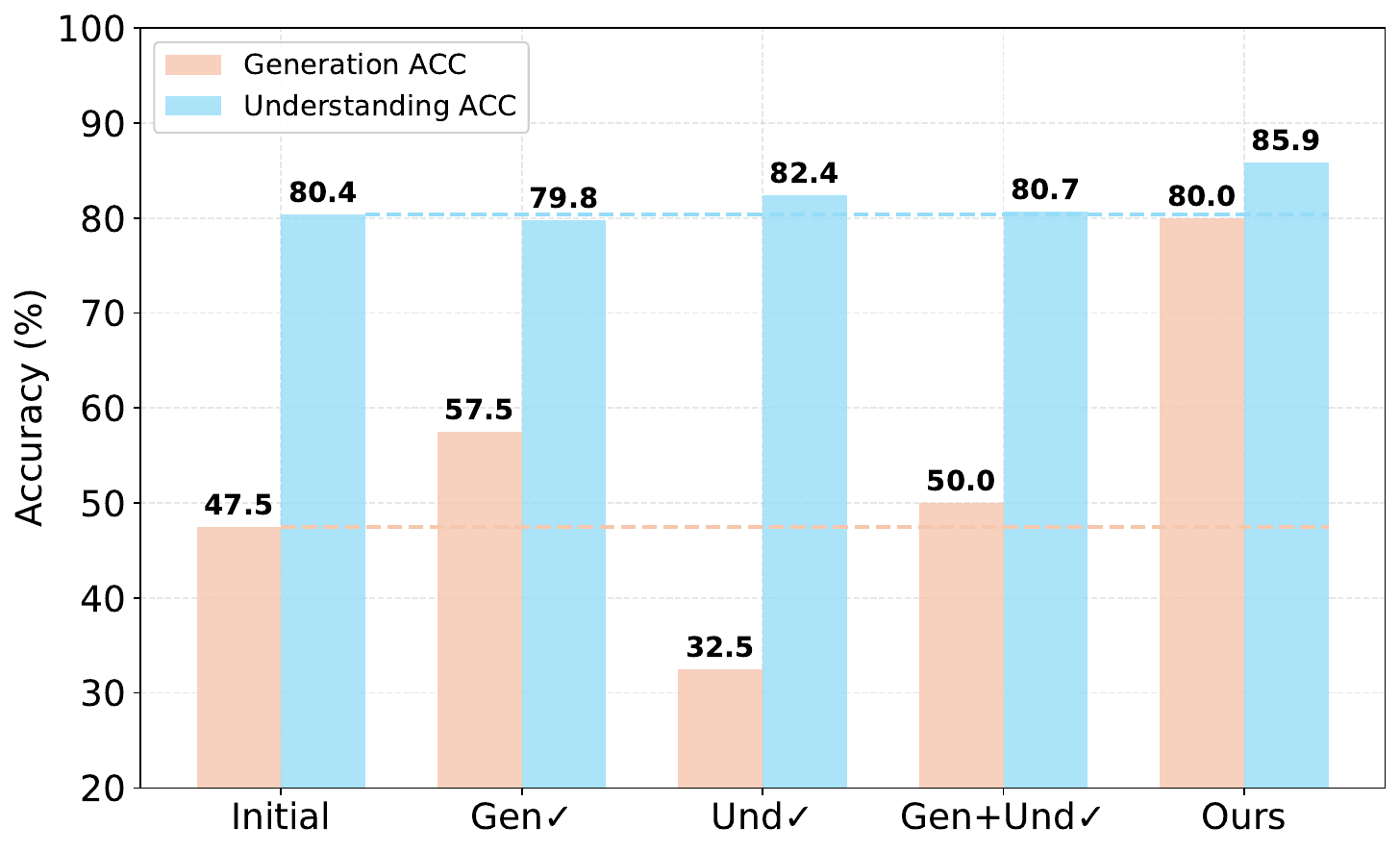}
    \caption{Fine-tuning BAGEL exclusively on generation or understanding degrades the complementary capability. Naive co-training shows minor gains, whereas our proposed method demonstrates significant improvement in both. Results are reported on counting subset of GenEval++.}
    \label{fig:uni_task}
\end{wrapfigure}

Our experiments demonstrate that incorporating reflecting and refining into the generative process enables the model to effectively leverage its understanding capability, leading to substantial improvements across multiple text-to-image benchmarks. Moreover, embedding reflection within generation not only enhances output quality but also exercises the model’s understanding ability, thereby preventing the degradation typically caused by modeling capacity competition. On tasks aligned with generative content, we even observe notable gains in understanding—for instance, counting accuracy improves from 79.3 to 84.6. Overall, the proposed R3 framework reconciles the long-standing conflict between generation and understanding: it achieves stronger generation while simultaneously preserving, rather than diminishing the understanding ability. This establishes a promising path forward for the development of future unified multimodal models.

Our contributions can be summarized as follows:
\begin{itemize}[leftmargin=*]

\item We provide a systematic analysis of the conflict between generation and understanding in multimodal large models, and identify its root cause: traditional approaches treat the two as independent tasks, which compete for model capacity and lead to a trade-off where one improves at the expense of the other.

\item Building on this insight, we propose the Reason–Reflect–Refine (R3) framework, which decomposes generation into a structured generate–understand–regenerate process. By explicitly incorporating understanding into the generative pipeline, R3 mitigates the conflict caused by separate optimization, yielding stronger generation while preserving understanding ability.

\item Extensive experiments verify the effectiveness of R3: by fully leveraging the model’s own understanding ability during generation, R3 not only achieves superior generative performance but also avoids the degradation of understanding. These findings shed light on the design of future unified multimodal models and point to new data strategies for balancing generation and understanding.

\end{itemize}

\section{Methodology}
\label{sec:methodology}

\subsection{Unifying Generation and Understanding}

\begin{figure}[t]
    \centering
    \includegraphics[width=0.9\linewidth]{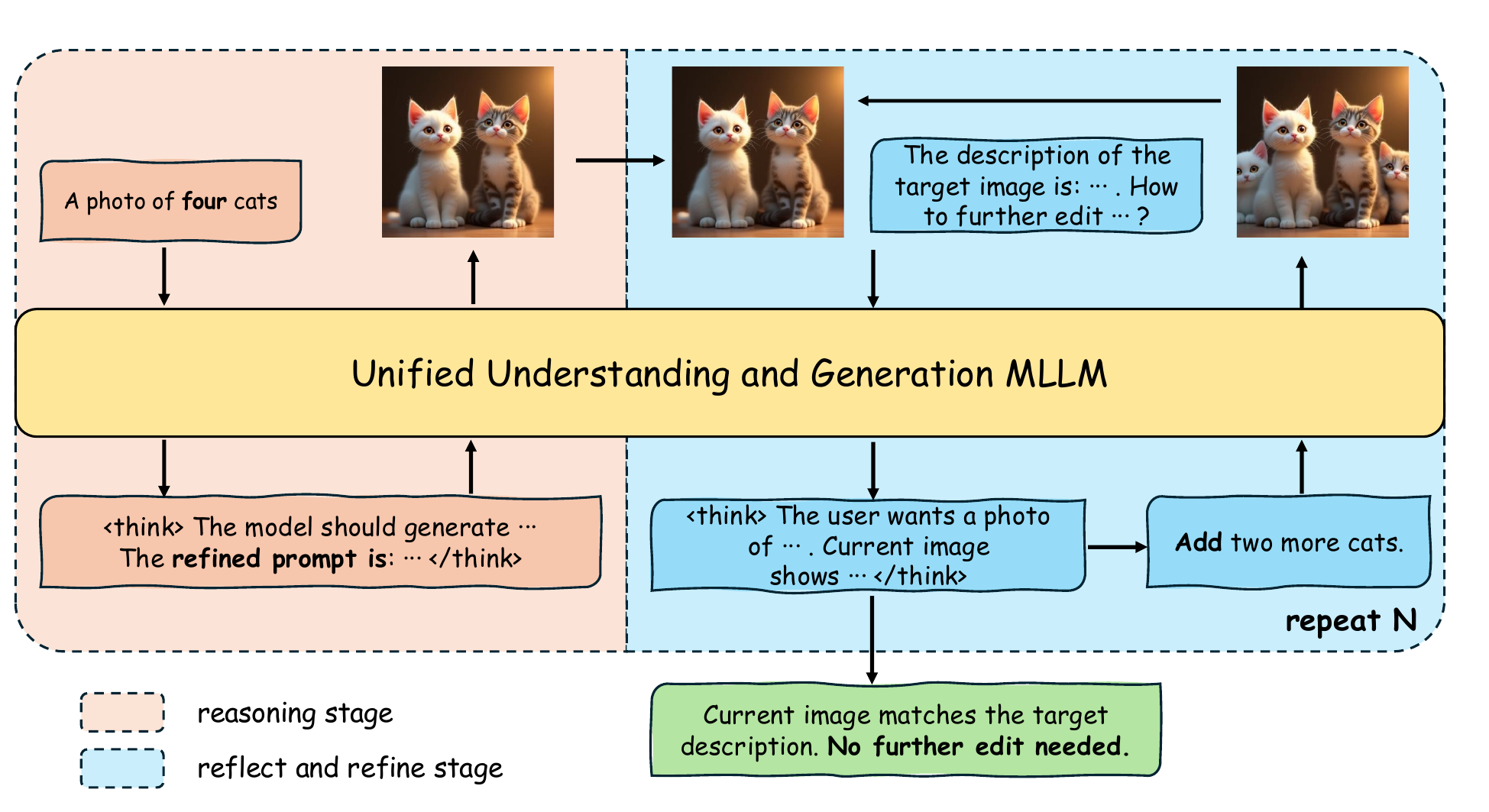}
    \caption{The inference pipeline of our Reason-Reflect-Refine framework. The model starts by Reasoning to produce an initial plan and image. It then enters an iterative Reflect-Refine loop, assessing its output and making corrections until the image aligns with the user's prompt or a stopping condition is met.}
    \label{fig:infer_pipeline}
    \vspace{-5pt}
\end{figure}

A key challenge in developing powerful, unified multi-modal models lies in effectively integrating both generation and understanding capabilities. As shown in~\cref{fig:uni_task}, fine-tuning a model exclusively on one of these tasks degrades its performance on the complementary task. This suggests a competitive relationship between the two tasks, where shared model parameters are optimized for one objective at the expense of the other. Furthermore, naive training on a mixture of data provides only negligible performance gains. This observation supports our core hypothesis: the optimization trajectories for generation and understanding are not inherently aligned. The two tasks are not coupled tightly enough to benefit from mutual training, suggesting they are driven by fundamentally different optimization dynamics. This observation leads us to a crucial question: Is it possible to align the optimization goals of generation and understanding, thereby reducing the conflicts that emerge when training a single model on both?

A possible solution is to intrinsically embedding visual understanding into the generative process. This paradigm shift ensures that the improvement of a model's generation capability is inherently dependent on its understanding ability, thereby preventing the common issue where training for one task degrades the performance of the other. To realize this vision, we introduce the Reason-Reflect-Refine (R3) framework(~\cref{fig:infer_pipeline}), a novel approach that recasts image generation as an iterative process of reasoning, reflection, and refinement. By embedding image understanding as a core component of a generative chain-of-thought, we enable the model to critically assess and progressively improve its own output, fostering a synergistic relationship between understanding and generation.

\subsection{Framework Overview}
\label{sec:framework_overview}

We build our framework on top of a unified multimodal model, BAGEL~\citep{bagel} (parametrized by $\theta$), which is capable of image understanding, generation, and editing. The generation process is formulated as a sequence of alternating text and image generation steps:
\begin{equation}
t^1,I^1, \dots, t^n, I^n \sim \pi_{\theta}(\cdot | c)
\end{equation}
where $c$ is the initial user-provided prompt, and at each step, the model generates text $t^i $ auto-regressively or an image $I^i$ by progressive denoising.

To make this process computationally tractable and modular, we decompose it into three distinct, alternating tasks. Assuming a Markovian property, we model the generation as a sequence of these specialized tasks:

\begin{enumerate}[leftmargin=*]
    \item \textbf{Reason}: The process begins with the model expanding on the input prompt $c$ to generate a more detailed, reasoned plan $t^1$. The plan is expected in the format \emph{"\textless{}think\textgreater{}plan\textless{}/think\textgreater{}"}. Then, the model synthesizes an initial image $I^1$ according to this plan. This is modeled by the joint probability $\pi_{\theta}(I^1,t^1 | c) = \pi_{\theta}(I^1 | t^1, c) \pi_{\theta}(t^1 | c)$. While $\pi_{\theta}(t^1 | c)$ is a standard language modeling policy, and the diffusion policy $\pi_{\theta}(I^1 | t^1, c)$ can be calculated using stochastic differential equations (SDEs), similar to~\citep{sit, flowgrpo}.

    \item \textbf{Reflect}: Upon obtaining an initial generated image, the model is then required to assess its alignment with the user's original intent c. This critical self-assessment process is termed reflection, which can be formally expressed as $\pi_{\theta}(t^{i+1} |I^i, c)$. In cases where the generated output is deemed satisfactory, the model is designed to produce a definitive termination signal: "No further edit needed." However, if the output is still deficient, the model performs a critical introspection, identifying the discrepancies between the current image and the desired objective. This introspection culminates in the generation of a refined editing instruction $e^{i+1}$, which serves as guidance for subsequent iterative improvements. To ensure a structured and consistent output, we employ a system prompt that strictly enforces the format: \emph{"<think>reflection</think>editing instruction."}

    \item \textbf{Refine}: Subsequently, the model executes the generated editing instruction $e^{i+1}$ to modify the previously created image $I^i$, and produce a refined output $I^{i+1}$. This refinement step is formally modeled as a conditional generation process:$\pi_{\theta}(I^{i+1} | e^{i+1}, I^i)$. The entire reflection-and-refinement loop is performed iteratively, forming a chain-of-thought that continues until the model's internal assessment confirms that the generated image satisfies all aspects of the user's request.
\end{enumerate}

\subsection{Tree-RL strategy}

\begin{figure}[t]
    \centering
    \includegraphics[width=0.9\linewidth]{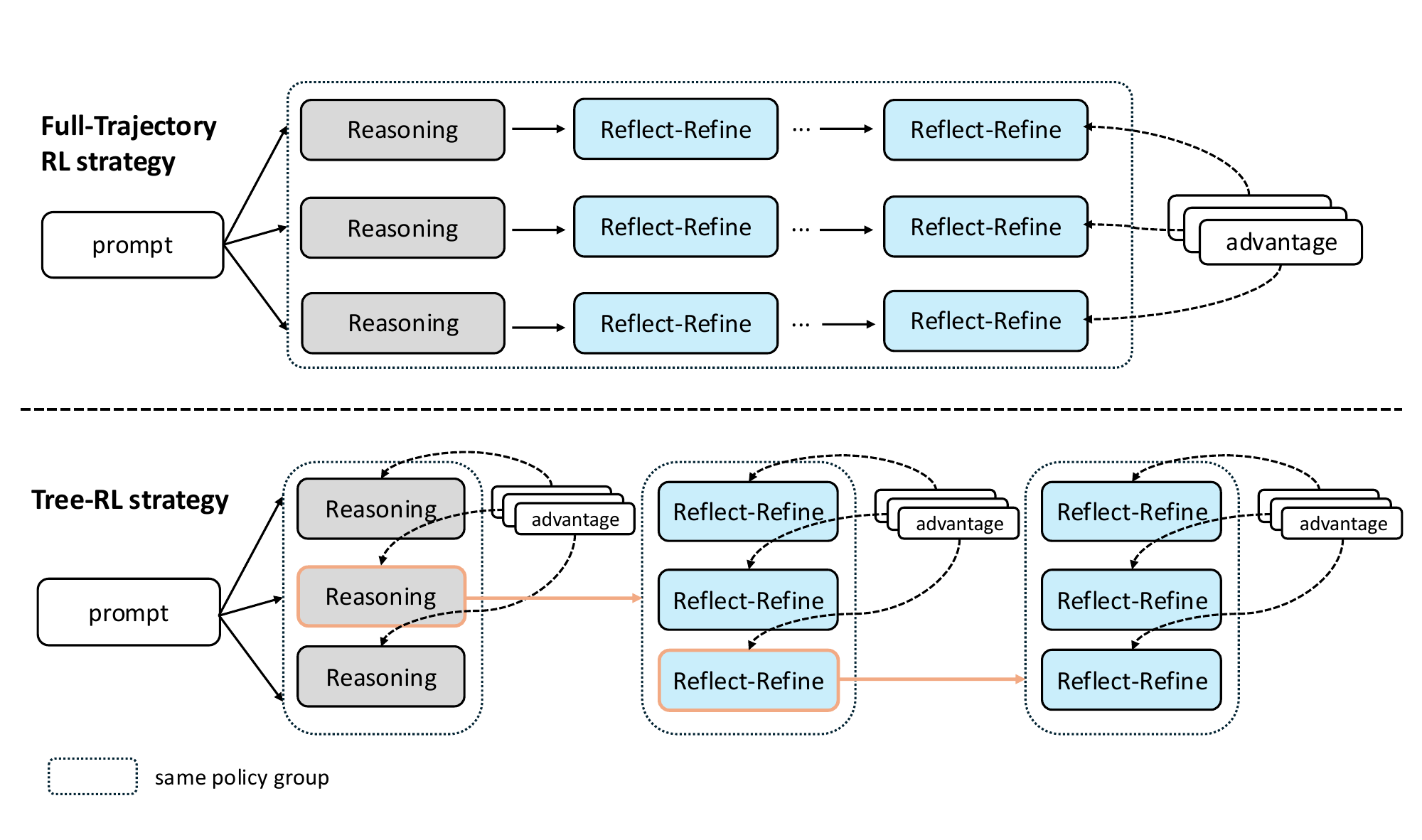}
    \caption{The training procedure, which alternates between optimizing the Reason policy and the Reflect-Refine policies. The replay buffer, populated by the Reason stage, provides on-policy data for training the subsequent stages.}
    \label{fig:train_pipeline}
    \vspace{-1em}
\end{figure}

The full, multi-turn trajectory can be conceptualized as a sequential process: \emph{Reason -> Reflect -> Refine -> Reflect -> Refine...}. This sequence can be theoretically framed as the chain-of-thought process of image generation, which could be directly optimized using reinforcement learning (RL). However, training end-to-end reinforcement learning (RL) models presents several challenges. First, the large number of training iterations can lead to error accumulation, causing instability. Second, the lack of explicit supervision for intermediate steps results in low training efficiency. To overcome these issues, we propose a tree-based RL strategy that provides clear supervision for the outcome of each intermediate step.

{As illustrated in~\cref{fig:train_pipeline}, we split the trajectory into Reason stage and Reflect-Refine stages. Each stage populates its result (result image $I^i$ and current reward) to the next stage as initial condition. To enhance training efficiency and speed up convergence, we employ an importance sampling strategy when selecting for previous stage's results: Sampling more samples with diverse rewards}. This strategic focus on error correction allows the model to learn and improve effectively, without compromising its overall task completion capability. All policies are optimized with the GRPO loss function, as described in~\cref{appendix:grpo}. As shown in~\cref{fig:tree_rl}, by employing this tree-based rollout strategy, our model can be trained effectively on long, complex generation chains, allowing for progressive improvement through multiple rounds of refinement during inference.

\begin{wrapfigure}[16]{r}{0.5\textwidth}
 \vspace{-2em}
    \centering
    \includegraphics[width=0.95\linewidth]{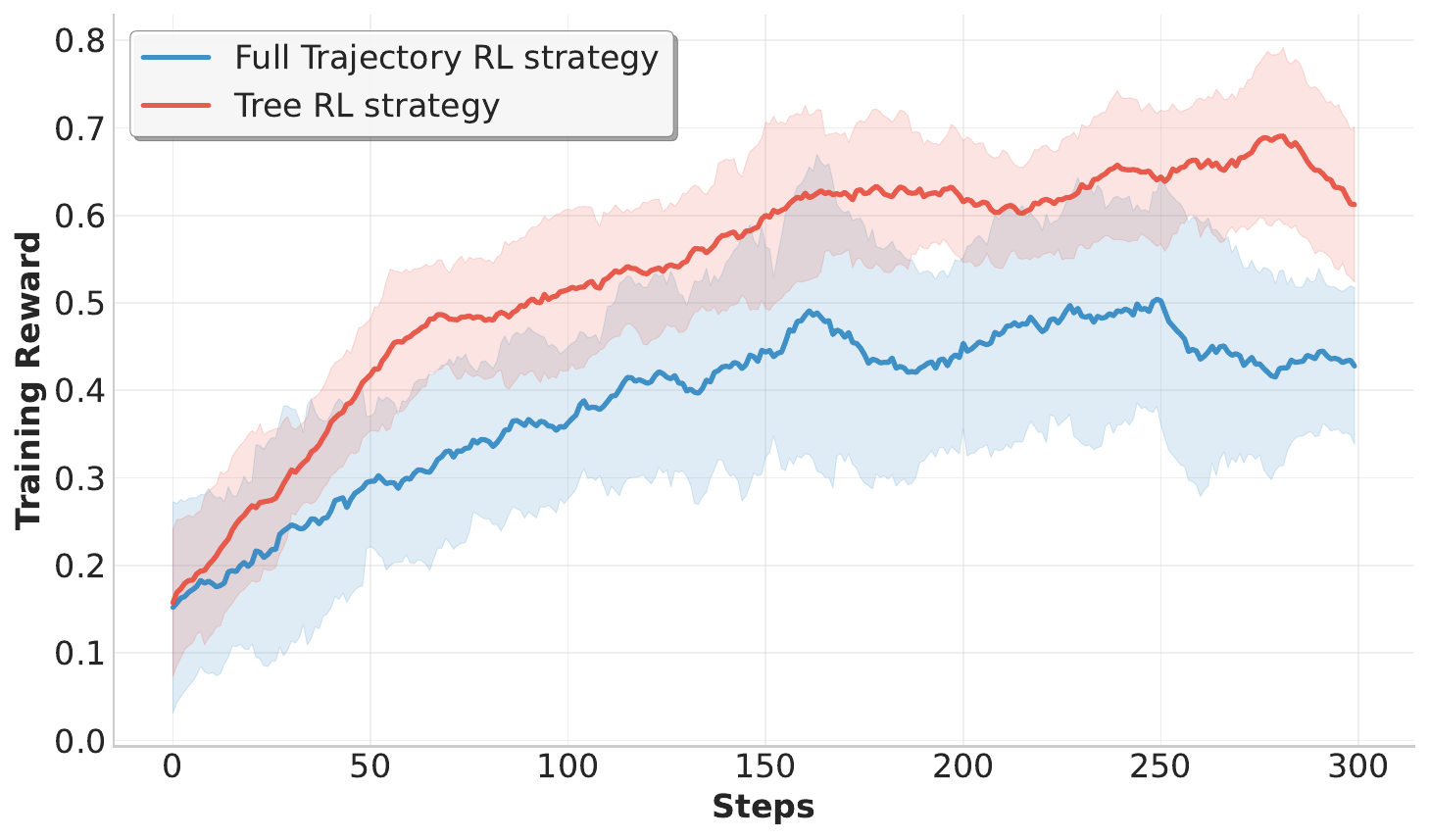}
    \caption{Training reward curves of the Tree-RL versus Full Trajectory RL strategies. The reward curve for Full Trajectory RL is substantially lower than that of Tree-RL. This performance gap is attributed to the high variance and noise introduced by the long trajectories inherent in the full trajectory approach, which complicates the advantage assignment problem.}
    \label{fig:tree_rl}
\end{wrapfigure}

\subsection{Stage-wise Reward}

We employ different reward models to evaluate the various stages of our reinforcement learning (RL) training. Specifically, as described in~\cref{sec:framework_overview}, the reasoning stage involves two policy processes: $\pi_{\theta}(I^1 | t^1, c)$ and $\pi_{\theta}(t^1 | c)$.

First, we measure the quality of the initial generated image $I^1$, using a pre-defined reward model $\mathcal{V}$ (a pre-trained Vision-Language Model). This model outputs a scalar score $V_j= \mathcal{V}(I^1_j, c) \in [0, 1]$, which represents the alignment between the image and the prompt. We then define the rewards for each policy as follows:
\begin{itemize}[leftmargin=*]

\item For the diffusion generation stage $\pi_{\theta}(I^1 | t^1, c)$, the reward is defined as $r_{j, \text{diffusion}} = V_j$.

\item For the text generation stage $\pi_{\theta}(t^1 | c)$, the reward is $r_{j, \text{text}} = V_j + r_{j, \text{format}}$, where $r_{j, \text{format}}$ measures the accuracy of the generated format.
\end{itemize}

For the Reflect-Refine stage, the model generates a textual reflection based on the image $\hat{I}$ and its reward $\hat{V}$ from the previous step. This process leads to one of two distinct outcomes:

\begin{enumerate}[leftmargin=*]
    \item \textbf{Correction Required}: If the image does not align well with the prompt $c$ (i.e., $\hat{V} < 1$), the model outputs a precise editing instruction $e_j$. The Refine step then generates an improved image $I_j$, which is evaluated to get a new image reward $V_j$.
    \item \textbf{Correction Unnecessary}: If the image already satisfies the prompt (i.e., $\hat{V} = 1$), the model must output the specific phrase "No further edit needed." This serves as a termination signal for the process.
\end{enumerate}

To encourage meaningful reflection and accurate termination, we use a correctness metric to evaluate the model's output. For an image with a reward of $\hat{V}$, the metric is defined as:

\begin{equation}
\label{eq:correctness}
\text{C}_j = 
\begin{cases} 
V_j - \hat{V} & \text{if } \hat{V} < 1 \\ 
\mathbb{I}(e_j = \text{"No further edit needed"}) & \text{if } \hat{V} = 1 
\end{cases}
\end{equation}
where $\mathbb{I}(\cdot)$ is the indicator function. This metric is designed to reward two key behaviors: measurable improvement for flawed images (where $V_j > \hat{V}$), and correct termination for images that already meet the prompt's criteria.

The rewards for the reflection and refinement steps are then based on this metric:
\begin{align}
r_{j, \text{reflection}} &= \text{C}_j + r_{j, \text{format}} \label{eq:reward_text_reflect}, \quad
r_{j, \text{refinement}} = \text{C}_j
\end{align}

While our reinforcement learning objective does not directly optimize understanding tasks, the model develops robust visual comprehension by being trained to evaluate image-prompt alignment via the reflection reward. 

\section{Experiments}
\label{sec:experiments}
\subsection{Co-evolution of Understanding and Generation}

To validate the co-evolution of understanding and generation capabilities within our framework, we conduct controlled experiments on the GenEval++ benchmark~\citep{echo4o}. 

\textbf{Experimental Setup:} Our training prompts are generated by randomly combining elements from the official templates, ensuring no overlap with the test set. We employ Qwen-2.5-VL-72B as our reward model. By default, our method includes an initial reasoning stage followed by four reflection-refinement stages. Following the original benchmark setting, we use GPT-4.1 to evaluate generation quality.

\textbf{Evaluation of Understanding:} To assess the model's understanding capabilities, we introduce two novel evaluation protocols: \textbf{Compositional Visual Question Answering (VQA)} and \textbf{Image-Text Alignment (ITA)}. The VQA task probes the model's ability to perceive compositional elements in its generated images. The ITA task evaluates its capacity to assess the overall quality and prompt-alignment of the generated images. 

For both tasks, we followed a standardized protocol. We first generated a corpus of images using a powerful, pretrained text-to-image generation model. Subsequently, we used the Gemini-2.5-Flash model to create the gold-standard ground truth for both VQA questions and ITA assessments and conduct a careful human alignment check to ensure the quality of annotation. Finally, we evaluated our model’s performance by calculating the similarity between its output and the annotated ground truth labels. This rigorous evaluation framework allows us to directly measure the effectiveness of our approach. We provide detailed description in ~\cref{sec:appendix_exp_details}. 

\begin{table}[h]
    \centering{
    \renewcommand{\arraystretch}{1.3}
    \caption{Instruction-following generation ability on the GenEval++ benchmark, evaluated by GPT-4.1. Bold indicates the best result. \dag indicates our framework with only the reasoning stage. Green arrows indicate improvement over the BAGEL baseline.}
    \resizebox{0.95\linewidth}{!}{
        \begin{tabular}{ccccccccc}
            \toprule
            Method & \multicolumn{1}{c}{Color} & \multicolumn{1}{c}{Count} & \multicolumn{1}{c}{Color/Count} & \multicolumn{1}{c}{Color/Pos} & \multicolumn{1}{c}{Pos/Count} & \multicolumn{1}{c}{Pos/Size} & \multicolumn{1}{c}{Multi-Count} & \multicolumn{1}{c}{Overall}\\
            \midrule
            \color[gray]{0.4}GPT-4o~\citep{gpt4o} & \color[gray]{0.4}0.900 & \color[gray]{0.4}0.675 & \color[gray]{0.4}0.725 & \color[gray]{0.4}0.625 & \color[gray]{0.4}0.600 & \color[gray]{0.4}0.800 & \color[gray]{0.4}0.850 & \color[gray]{0.4}0.739 \\
            \midrule
            FLUX.1-Kontext~\citep{FLUX} & 0.425 & 0.500 & 0.200 & 0.250 & 0.300 & 0.400 & 0.325 & 0.343 \\
            FLUX.1-dev~\citep{FLUX} & 0.350 & 0.625 & 0.150 & 0.275 & 0.200 & 0.375 & 0.225 & 0.314 \\
            Janus-Pro~\citep{januspro} & 0.450 & 0.300 & 0.125 & 0.300 & 0.075 & 0.350 & 0.125 & 0.246\\
            T2I-R1~\citep{t2ir1} & 0.675 & 0.325 & 0.200 & 0.350 & 0.075 & 0.250 & 0.300 & 0.311 \\
            Echo-4o~\citep{echo4o} & \textbf{0.800} & 0.575 & 0.550 & \textbf{0.775} & 0.625 & \textbf{0.800} & 0.625 & 0.679 \\
            \midrule
            BAGEL~\citep{bagel} & 0.325 & 0.600 & 0.250 & 0.325 & 0.250 & 0.475 & 0.375 & {0.371} \\
            BAGEL + Ours\dag & 0.500 & 0.650 & 0.600 & 0.650 & 0.550 & 0.600 & 0.600 & {0.593} \rlap{\ \textcolor{green!60!black}{\small$\uparrow$0.22}} \\
            BAGEL + Ours & 0.675 & \textbf{0.725} & \textbf{0.575} & 0.725 & \textbf{0.750} & 0.575 & \textbf{0.800} & {\textbf{0.689}} \rlap{\ \textcolor{green!60!black}{\small$\uparrow$0.32}}  \\
            \bottomrule
        \end{tabular}
    }

    \label{tab:overall_gen}
    \vspace{-1em}
    }
\end{table}

\begin{table}[htbp]
    \centering{
    \renewcommand{\arraystretch}{1.3}
    \caption{Evaluation of understanding capabilities on our proposed ITA benchmarks. All scores are reported as accuracy (\%). \dag indicates our framework with only the reasoning stage. Green arrows indicate improvement over the BAGEL baseline.}
    \resizebox{0.95\linewidth}{!}{
        \begin{tabular}{ccccccccc}
            \toprule
            ITA & \multicolumn{1}{c}{Color} & \multicolumn{1}{c}{Count} & \multicolumn{1}{c}{Color/Count} & \multicolumn{1}{c}{Color/Pos} & \multicolumn{1}{c}{Pos/Count} & \multicolumn{1}{c}{Pos/Size} & \multicolumn{1}{c}{Multi-Count} & \multicolumn{1}{c}{Overall}\\
            \midrule
            BAGEL & 60.63 & 58.54 & 45.42 & 63.54 & 53.96 & 80.83 & 61.50 & 60.60 \\
            BAGEL + Ours\dag  & 60.42 & 59.38 & 47.71 & 63.75 & 55.63 & 81.46 & 63.96 & 61.76 \rlap{\ \textcolor{green!60!black}{\small$\uparrow$1.16}} \\
            BAGEL + Ours & \textbf{69.58} & \textbf{67.50} & \textbf{69.79} & \textbf{72.29} & \textbf{76.04} & \textbf{83.33} & \textbf{75.00} & \textbf{73.37} \rlap{\ \textcolor{green!60!black}{\small$\uparrow$12.77}} \\
            \bottomrule
        \end{tabular}
    }
    \label{tab:itaa_results}
    }
\end{table}

\begin{table}[htbp]
    \centering{
    \renewcommand{\arraystretch}{1.3}
    \caption{Evaluation of VQA capabilities. All scores are reported as accuracy (\%). \dag indicates our framework with only the reasoning stage. Green arrows indicate improvement over the BAGEL baseline.}
    \resizebox{0.95\linewidth}{!}{
        \begin{tabular}{ccccccccc}
            \toprule
            VQA & \multicolumn{1}{c}{Color} & \multicolumn{1}{c}{Count} & \multicolumn{1}{c}{Color/Count} & \multicolumn{1}{c}{Color/Pos} & \multicolumn{1}{c}{Pos/Count} & \multicolumn{1}{c}{Pos/Size} & \multicolumn{1}{c}{Multi-Count} & \multicolumn{1}{c}{Overall}\\
            \midrule
            BAGEL & 91.74 & 79.30 & 88.28 & 77.99 & 82.93 & 85.10 & 92.45 & 86.48 \\
            BAGEL + Ours\dag  & 91.67 & 76.12 & 88.76 & 78.71 & 83.29 & 84.98 & 93.45 & 86.72 \rlap{\ \textcolor{green!60!black}{\small$\uparrow$0.24}} \\
            BAGEL + Ours & \textbf{93.95} & \textbf{84.63} & \textbf{91.15} & \textbf{84.09} & \textbf{86.06} & \textbf{86.54} & \textbf{94.50} & \textbf{89.63} \rlap{\ \textcolor{green!60!black}{\small$\uparrow$3.15}} \\
            \bottomrule
        \end{tabular}
    }
    \label{tab:overall_und}
    }
\end{table}

\begin{figure}[t]
    \centering
    \includegraphics[width=0.99\linewidth]{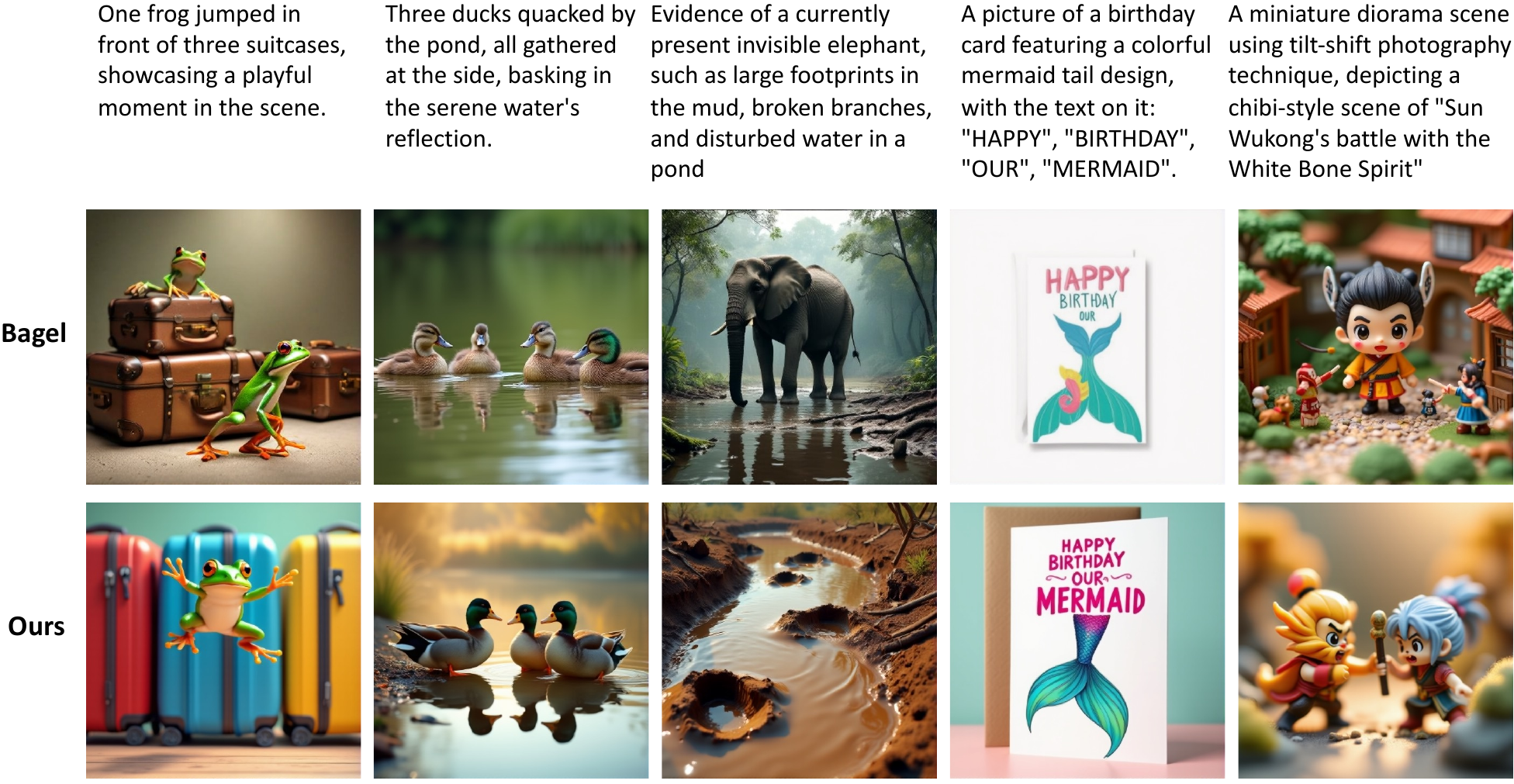}
    \caption{Qualitative comparison between Bagel and our results.}
    \label{fig:vis}
    \vspace{-1em}
\end{figure}

\begin{table}[htbp]
    \begin{minipage}[t]{0.4\linewidth}
        \centering
        \caption{Impact of training trajectory length on generation (GenEval++ Validation Reward) and understanding (ITA). "RR" denotes a Reflection-Refinement stage.}
        \begin{tabular}{ccc}
        \toprule
          Rollout Stage   &  GenEval++ & ITA\\
          \midrule
           Reason & 0.654  & 62.83 \\
           \midrule
           Reason + $1\times$ RR  & 0.729 & 74.49 \\
           Reason + $2\times$ RR  & 0.732 & 74.76 \\
             \bottomrule
        \end{tabular}
        \label{tab:multi_turn_training}
    \end{minipage}\hfill
    \begin{minipage}[t]{0.55\linewidth}
            \centering
        \caption{Cross-topic evaluation on GenEval++. Models are trained on specific category and tested on all categories.}
        \begin{tabular}{cccc}
        \toprule
        \diagbox{Train}{Test} & Counting & Color & Position/Size \\
        \midrule
        \textcolor{gray}{Pretrained} &\textcolor{gray}{62.71} & \textcolor{gray}{60.63} & \textcolor{gray}{80.83} \\
        \midrule
        Counting & \textbf{71.25} & 60.63 & 81.46 \\
        Color & 62.50 & \textbf{66.25} & 81.25  \\
        Position/Size & 60.00 & 60.00 & \textbf{83.33} \\
        
        \bottomrule
        \end{tabular}
        \label{tab:understand_generalize}
    \end{minipage}
    \vspace{-1em}
\end{table}

\begin{figure}[t]
    \centering
    \includegraphics[width=0.99\linewidth,height=0.18\textheight]{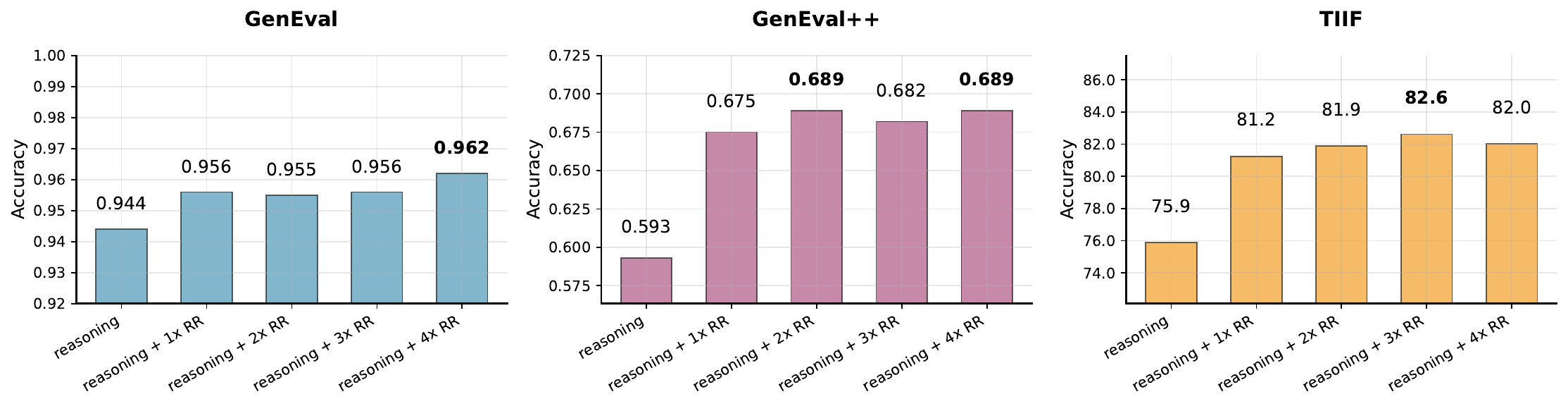}
    \caption{Inference-time scaling effect across the GenEval, GenEval++, and TIIF benchmarks (left to right). Performance is shown as a function of the maximum allowed reflection-refinement turns.}
    \label{fig:infer_scaling}
    \vspace{-1em}
\end{figure}
\begin{figure}[t]
    \centering
    \includegraphics[width=0.9\linewidth,height=0.18\textheight]{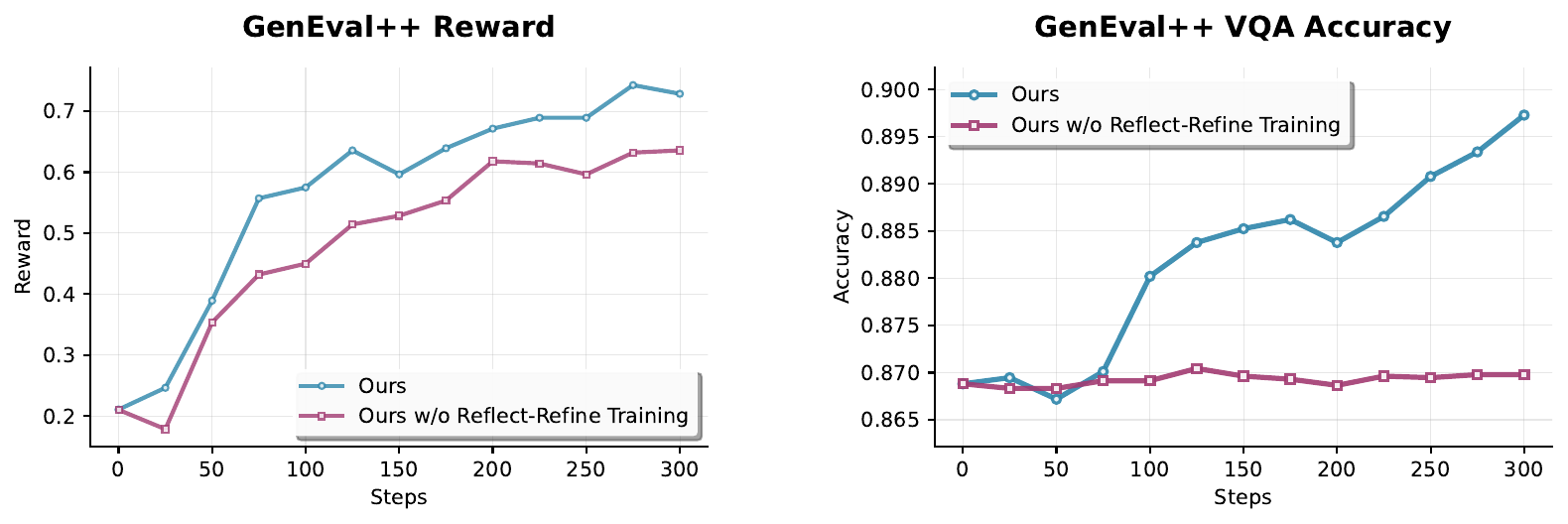}
    \caption{The evolution of generation and understanding abilities in the training process. The left figure shows the generation accuracy (measured by Qwen-2.5-VL-72B.) and the right figure shows the VQA accuracy. }
    \label{fig:evolution_gen_und}
    \vspace{-2em}
\end{figure}

The results in~\cref{tab:overall_gen} confirm the enhanced generation capabilities of our framework. Even compared with the SOTA method Echo-4o~\citep{echo4o} which finetunes on related data, our method still achieves 1\% point overall improvement. And in complex situation like \emph{Multi-Count}, our method is significantly better then Echo-4o.  

As shown in~\cref{tab:itaa_results} and~\cref{tab:overall_und}, our approach also yields substantial gains in the ITA and VQA understanding tasks. Notably, the reflection-refinement process is critical. While a baseline with only the reasoning stage improves GenEval++ scores, our full framework adds a further improvement of nearly 10\% points. This effect is more dramatic in understanding tasks, where the reasoning-only baseline offers minimal gains (1.16 on ITA, 0.24 on VQA), whereas our full framework achieves significant improvements (12.77 on ITA, 3.15 on VQA). These findings highlight that the reflection-refinement stage is essential for improving generation and is the key to unlocking the model's understanding abilities. We also provide some qualitative results in~\cref{fig:vis}. For a more detailed visualization of the multi-round generation process, please refer to the ~\cref{app:mr_vis}.

\subsection{Ablation Studies}

\noindent\textbf{Influence of Trajectory Length.} 
We investigate the impact of trajectory length on performance during both training and inference. For training, as detailed in ~\cref{tab:multi_turn_training}, a trajectory length of two (one reasoning stage and one reflection-refinement stage) achieves an optimal balance between computational cost and performance.

For our inference-time analysis, we evaluate the impact of trajectory length by setting a maximum number of reflection-refinement turns. For each sample, the evaluation terminates in one of two ways: either the model signals completion by outputting "No further edit needed," or it is stopped upon reaching the predefined maximum number of turns. As illustrated in ~\cref{fig:infer_scaling}, the results across the GenEval~\citep{geneval}, GenEval++~\citep{echo4o}, and TIIF~\citep{tiif} benchmarks show a consistent pattern. The most substantial performance gain occurs after the first reflection-refinement turn. While further improvements are observed with additional turns, the performance generally saturates, reaching its peak when the trajectory length is extended to four or five turns.

\noindent\textbf{Capabilities Evolution During Training.}
~\cref{fig:evolution_gen_und} demonstrates how our proposed framework drives the simultaneous improvement of generation and understanding capabilities. In the first 150 steps, our model's generation accuracy follows a trajectory similar to the reflection-free baseline, while the VQA accuracy shows minimal change. This suggests that the initial training primarily focuses on basic generative mapping without deepening internal understanding. However, beyond the 150-step mark, the reflection-refinement mechanism begins to yield observable results. At this point, the model's understanding ability starts to rise noticeably. This gain in understanding does not occur in isolation; it is directly associated with a subsequent acceleration in generation accuracy, allowing our model to achieve superior performance relative to the baseline. Ultimately, the comprehensive results validate the effectiveness of our framework in mitigating the optimization dilemma and demonstrating that explicitly integrated understanding is key to unlocking superior, unified multimodal performance.

\noindent\textbf{Generalization of co-evolutionary Learning.}
A key question is whether the co-evolution of understanding and generation capabilities generalizes to broader domains beyond GenEval++. While rigorously measuring this co-evolution would require specialized evaluation protocols, as we constructed for GenEval++, we can assess the impact on general generation capabilities. Our results on the TIIF benchmark (~\cref{tab:tiif_qwen}) demonstrate that our framework yields significant improvements in a general-domain setting, suggesting that the benefits of our approach do transfer.

To further probe the nature of the learned understanding, we conducted a cross-topic experiment within GenEval++. We trained our model on specific attribute categories (e.g., color, counting) and evaluated its understanding on all categories. The results, presented in ~\cref{tab:understand_generalize}, indicate that the improvements in understanding are localized to the domains the model was trained on. This finding suggests that our current framework learns domain-specific understanding. While this specialized knowledge is highly effective, developing methods to foster more generalized understanding remains an important direction for future research.

\begin{table}[t]
 \centering
 \caption{\textbf{Quantitative evaluation results on TIIF testmini benchmark~\citep{tiif}.} Results are evaluated by QwenVL2.5-72B. * indicates our reproduced results using the official repository and checkpoint. The best results are \textbf{bolded}.}
 \renewcommand{\arraystretch}{1.7} 
 \setlength{\tabcolsep}{3pt}
 \centering
 \resizebox{0.99\linewidth}{!}{
 \begin{tabular}{ccccccccccccc}
 \toprule
 \multirow{2}{*}{\textbf{Model}}
   & \multirow{2}{*}{\textbf{Overall}}
   & \multicolumn{4}{c}{\textbf{Basic Following}}
   & \multicolumn{6}{c}{\textbf{Advanced Following}}
   & \multirow{2}{*}{\textbf{Designer}} \\

 \cmidrule(lr){3-6} \cmidrule(lr){7-12}

 & &
   Avg & Attribute & Relation & Reasoning &
   Avg & \makecell{Attribute\\+Relation} & \makecell{Attribute\\+Reasoning} & \makecell{Relation\\+Reasoning} & Style & Text & \\

 \midrule
\color[gray]{0.4}GPT-4o~\citep{gpt4o} &\color[gray]{0.4}84.19	&\color[gray]{0.4}85.30	&\color[gray]{0.4}81.00	&\color[gray]{0.4}86.16	&\color[gray]{0.4}88.74	&\color[gray]{0.4}81.24	&\color[gray]{0.4}81.95	&\color[gray]{0.4}80.03	&\color[gray]{0.4}80.88	&\color[gray]{0.4}76.67	&\color[gray]{0.4}92.76	&\color[gray]{0.4}89.55  \\
\midrule
FLUX.1-dev~\citep{FLUX}  &66.24	&74.41	&72.50	&78.20	&72.52	&60.72	&66.76	&61.76	&56.60	&63.33	&44.49	&74.63 \\
FLUX.1-Pro~\citep{FLUX} &63.75	&71.39	&70.00	&68.51	&75.66	&64.63	&70.69	&62.34	&64.65	&63.00	&34.39	&69.94 \\
 Janus-Pro-7B~\citep{januspro} &65.38	&74.99	&74.50	&73.69	&76.77	&61.77	&65.71	&62.01	&61.16	&43.33	&38.46	&79.48 \\
 T2I-R1~\citep{t2ir1} &67.61	&81.14	&80.50	&\textbf{83.09}	&79.81	&67.38	&69.92	&70.10	&68.69	&50.00	&32.13	&74.25 \\
 \midrule
 BAGEL~\citep{bagel}* &70.97  &78.16  &78.00  &80.24  &76.25  &68.23  &73.37  &64.36  &68.92  &80.00  &40.72  &76.87  \\
 BAGEL $w/$ self-CoT~\citep{bagel}* &68.06 &77.63  &75.00  &78.55  &79.33  &71.24  &\textbf{77.65}  &69.77  &72.93  &69.93  &26.24   &69.78  \\
 BAGEL + Ours & \textbf{82.02} & \textbf{85.07} & \textbf{84.00} & 82.61 & \textbf{88.58} & \textbf{78.52} & 76.96 & \textbf{80.07} & \textbf{77.43} & \textbf{93.33} & \textbf{72.40} & \textbf{82.84} \\
\bottomrule
\end{tabular}
}
\vspace{-1em}
\label{tab:tiif_qwen}
\end{table}



\section{Related Work}
\label{related_work}
\textbf{Unified Large Multi-modal Models.} Unifying and utilizing multi-modal representations is a central goal in the development of large language models. Early work, such as Chameleon~\citep{team2024chameleon}, unified image and text by utilizing discrete tokens for both generation and understanding, employing a unified next-token prediction paradigm. Later works~\citep{janus,januspro,vilau,unitok,unilip,uniworld,emu3} extended this paradigm with improved tokenizers, decoupled encoders for generation and understanding, and even lossless continuous tokens. In contrast, \citet{wu2024next} unifies the representation of different modalities in the latent space, utilizing different encoders and decoders for each modality. More recent work~\citep{transfusion,showo,showo2,bagel} has further bridged the modality gap by employing different tasks within the same model: next-token prediction for understanding and a diffusion objective for generation. Among these, BAGEL~\citep{bagel} has emerged as a particularly powerful model, demonstrating significantly improved performance on both generation and understanding tasks. In this paper, we also focus on unified multi-modal modeling, but from a task-oriented perspective. Specifically, we aim to unify tasks by framing understanding as a subtask of generation and thereby improving the generation and understanding simultaneously ~\citep{an2025unictokens}. We adopt BAGEL as our baseline model to ensure a convincing evaluation.

\textbf{Reinforcement Learning for Multi-modal Models.} Reinforcement Learning (RL) has emerged as a powerful paradigm for enhancing the reasoning capabilities of large language models~\citep{openaio1,deepseekmath}. This approach enables models to transcend the limitations of imitation learning, where they merely replicate demonstrated patterns. Instead, RL empowers them to autonomously discover and optimize complex, multi-step generative strategies. Early works~\cite{team2025kimi,zhou2025r1} aimed to adopt the RL paradigm to image understanding, which still focused on text level learning. Recent works have begun to leverage RL to improve text-to-image synthesis~\citep{t2ir1,reasongenr1,gotr1,flowgrpo}. For instance, GoT-R1~\citep{gotr1}, built on the next-token prediction paradigm, proposes learning a detailed semantic plan and layout for image generation. T2I-R1~\citep{t2ir1} further employs reinforcement learning to jointly train both text and image tokens. FlowGRPO~\citep{flowgrpo} extends this approach to diffusion models by applying Generative Reward Process Optimization (GRPO) to the series of denoising steps. Since our work builds on BAGEL, which generates text via discrete next-token prediction and images via a diffusion process, we adopt both GRPO and FlowGRPO to jointly optimize both processes. However, our focus is on recasting understanding as a component of generation. We therefore compose multiple tasks to construct the generation process, resulting in a more complex and thorough framework.

\section{Conclusion}
This paper analyzed the key trade-off between generation and understanding in multimodal models, identifying the conflict may arise from their competitive optimization objectives.
To navigate this challenge, we proposed the Reason-Reflect-Refine (R3) framework. R3 re-conceptualizes generation as a multi-step process where the model explicitly leverages its understanding capability to iteratively refine its output. Our experiments indicated that by integrating reflection into the generative flow, R3 helps to ease the optimization dilemma, leading to stronger generation results and improved understanding ability related to the generative task. This framework offers valuable insights for designing next-generation unified multimodal models.
\newpage

\section*{Acknowledgments}
LW is supported by National Science and Technology Major Project (2022ZD0114902) and National
Science Foundation of China (NSFC92470123, NSFC62276005).



\bibliography{iclr2026_conference}
\bibliographystyle{iclr2026_conference}
\newpage
\appendix
\section{Appendix}
\subsection{Clarification on the Use of LLM}
During the preparation of this manuscript, Large Language Model (LLM) was utilized as a tool to enhance readability and correct grammatical errors. The authors carefully reviewed and edited all AI-generated suggestions to ensure the final text aligns with their original intent and arguments. The intellectual content, analyses, and all arguments presented in this paper are solely the work of the human authors, who take full responsibility for the final content of the publication.

\subsection{RL Training with GRPO}
\label{appendix:grpo}
Our methodology leverages Group-Relative Policy Optimization (GRPO) to refine two key components: the Chain-of-Thought (CoT) generation within our Reasoning and Reflect stages, and the denoising process for image generation and editing. For the text-based CoT policy, we employ the standard GRPO algorithm~\citep{deepseekmath}. For the diffusion model, we adopt the framework of FlowGRPO~\citep{flowgrpo}. Below, we provide a technical overview of these reinforcement learning techniques.

\subsubsection{Group-Relative Policy Optimization (GRPO)}
Group-Relative Policy Optimization (GRPO)~\citep{deepseekmath} is a policy gradient algorithm derived from Proximal Policy Optimization (PPO). It stabilizes training by normalizing advantages against a group of responses sampled from the policy. This approach reduces the variance of advantage estimates, leading to more consistent and effective policy updates.

The optimization process begins by sampling a group of $G$ responses $\{o_i\}_{i=1}^G$ for a given prompt $q$ using the current policy $\pi_{\theta_{\mathrm{old}}}$. Each response $o_i$ is evaluated to obtain a reward $R_i$. The group-relative advantage $\hat{A}_{i,t}$ for the $i$-th response is then calculated by standardizing its reward against the statistics of the entire group:
$$
\hat{A}_{i,t} = \frac{R_i - \mathrm{mean}\bigl(\{R_j\}_{j=1}^G\bigr)}{\mathrm{std}\bigl(\{R_j\}_{j=1}^G\bigr) + \delta},
$$
where $\delta$ is a small constant added for numerical stability.

The policy $\pi_\theta$ is then updated by maximizing a clipped surrogate objective. This objective is regularized with a KL-divergence penalty to prevent large deviations from a reference policy $\pi_{\mathrm{ref}}$ (typically the initial supervised fine-tuned model):
$$
\begin{aligned}
\mathcal{J}_{\mathrm{GRPO}} & (\theta) ={} \mathbb{E}_{\substack{(q,a)\sim\mathcal{D}, \\ \{o_i\}_{i=1}^G\sim\pi_{\theta_{\mathrm{old}}}(\cdot\mid q)}} \\
& \left[ \frac{1}{\sum_{i=1}^G |o_i|}
\sum_{i=1}^G \sum_{t=1}^{|o_i|}
\min\bigl(r_{i,t}(\theta)\,\hat{A}_{i,t},\,
\mathrm{clip}\bigl(r_{i,t}(\theta),\,1-\varepsilon,\,1+\varepsilon\bigr)\,\hat{A}_{i,t}\bigr)
- \beta\,D_{\mathrm{KL}}\bigl(\pi_\theta\|\pi_{\mathrm{ref}}\bigr)
\right].
\end{aligned}
$$
Here, $r_{i,t}(\theta)$ is the per-token importance sampling ratio between the new policy $\pi_\theta$ and the old policy $\pi_{\theta_{\mathrm{old}}}$:
$$
r_{i,t}(\theta) = \frac{\pi_\theta\bigl(o_{i,t}\mid q,\,o_{i,<t}\bigr)}{\pi_{\theta_{\mathrm{old}}}\bigl(o_{i,t}\mid q,\,o_{i,<t}\bigr)}.
$$

\subsubsection{FlowGRPO for Diffusion Model Optimization}
To apply policy optimization to our diffusion model, we utilize FlowGRPO~\citep{flowgrpo}, which adapts the GRPO framework for continuous state-space models trained via flow matching.

\paragraph{Flow Matching Preliminaries.}
Our diffusion model is built upon Rectified Flow~\citep{flowmatching}, which defines a linear interpolation between a data sample $x_0 \sim \mathcal{X}_0$ and a noise sample $x_1 \sim \mathcal{X}_1$ as $x_t = (1-t)x_0 + t x_1$ for $t \in [0,1]$. A velocity field network $v_\theta(x_t, t)$ is trained to predict the vector field $v = x_1 - x_0$ via the flow matching objective:
$$
\mathcal{L}_{\text{FM}}(\theta) = \mathbb{E}_{t,x_0,x_1} \left[\|(x_1 - x_0) - v_\theta((1-t)x_0 + t x_1, t)\|_2^2 \right].
$$

\paragraph{MDP Formulation and Policy Optimization.}
The iterative denoising process of a diffusion model can be naturally framed as a Markov Decision Process (MDP). At each discrete step $t$, the state is $s_t = (c, t, x_t)$, comprising the conditioning prompt $c$, the current time $t$, and the noisy sample $x_t$. The action $a_t$ corresponds to generating the subsequent, less noisy sample $x_{t-1} \sim \pi_\theta(x_{t-1}|x_t,c)$. A terminal reward $R(x_0, c)$ is assigned only at the final step ($t=0$) upon generating the complete sample $x_0$.

Under this MDP formulation, FlowGRPO generates a group of $G$ images $\{x_0^i\}_{i=1}^G$ for a given prompt $c$. The advantage for the $i$-th trajectory is computed relative to the group's final rewards:
$$
\hat{A}^i = \frac{R(x_0^i,c) - \mathrm{mean}(\{R(x_0^j,c)\}_{j=1}^G)}{\mathrm{std}(\{R(x_0^j,c)\}_{j=1}^G) + \delta}.
$$
Since the reward is terminal, the advantage $\hat{A}^i$ is constant across all timesteps $t$ for a given trajectory $i$. The FlowGRPO objective is then:
$$
\begin{aligned}
\mathcal{J}_{\text{Flow-GRPO}}(\theta) ={} & \mathbb{E}_{c, \{x^i\}_{i=1}^G} \\
& \left[
\frac{1}{G}\sum_{i=1}^G \frac{1}{T} \sum_{t=1}^{T}
\min\big(r_t^i(\theta) \hat{A}^i, \text{clip}(r_t^i(\theta),1-\epsilon,1+\epsilon)\hat{A}^i \big) - \beta D_{\text{KL}}(\pi_\theta \| \pi_{\text{ref}})
\right],
\end{aligned}
$$
where the importance ratio is $r_t^i(\theta) = \frac{p_\theta(x_{t-1}^i|x_t^i,c)}{p_{\theta_{\mathrm{old}}}(x_{t-1}^i|x_t^i,c)}$.

\paragraph{Stochastic Differential Equation for Exploration.}
Reinforcement learning requires policy exploration, yet the generative process defined by an Ordinary Differential Equation (ODE), $dx_t = v_\theta(x_t, t) dt$, is deterministic. To facilitate exploration, FlowGRPO converts the ODE into a corresponding Stochastic Differential Equation (SDE):
$$
dx_t = \left(v_\theta(x_t,t) + \frac{\sigma_t^2}{2t} (x_t + (1-t)v_\theta(x_t,t))\right) dt + \sigma_t dw_t,
$$
where $dw_t$ represents the increments of a Wiener process and $\sigma_t$ controls the noise magnitude. Discretizing this SDE with the Euler-Maruyama method yields the update rule for the generative sampling process:
$$
x_{t-\Delta t} = x_t - \left(v_\theta(x_t,t) + \frac{\sigma_t^2}{2t}(x_t + (1-t)v_\theta(x_t,t))\right)\Delta t + \sigma_t \sqrt{\Delta t} z, \quad z \sim \mathcal{N}(0,I).
$$
The noise schedule is given by $\sigma_t = a \sqrt{t/(1-t)}$, where $a$ is a scalar hyperparameter that adjusts the level of stochasticity. This SDE-based sampling enables the policy to explore different generation paths, which is essential for effective RL-based optimization.

\paragraph{Efficient Training with MixGRPO~\citep{mixgrpo}.} To mitigate the computational demands of SDE-based sampling during policy training, we adopt the MixGRPO~\citep{mixgrpo} strategy. This approach combines SDE and ODE sampling in denoising process, substantially reducing the computational burden while maintaining strong model performance.

\subsection{Experimental Details}
\label{sec:appendix_exp_details}

\subsubsection{Generation Benchmarks}
To validate the effectiveness of our proposed \textbf{Reason-Reflect-Refine} framework, we conducted experiments on three widely-used text-to-image generation benchmarks.

\paragraph{GenEval.} The GenEval benchmark~\citep{geneval} is designed to assess a model's capability to generate images with complex compositional requirements, including object counting, spatial relationships, and attribute binding. For training, prompts were generated using the official GenEval scripts, which construct prompts by randomly combining predefined templates. We rigorously excluded any prompts present in the official test set from our training data. The reward signal for our reinforcement learning process is derived from the soft reward function proposed by FlowGRPO~\citep{flowgrpo}, which was specifically designed for this benchmark. The result is presented in~\cref{tab:geneval}, our method reaches a new state-of-the-art with a score of 0.962.

\begin{table}[htbp]
\centering
\caption{Hyperparameters for Different Datasets}
\label{tab:hyperparams}
\begin{tabular}{l|ccc}
\hline
\textbf{Hyperparams} & \textbf{GenEval} & 

\textbf{GenEval++} & \textbf{TIIF} \\
\hline
Learning Rate & 5e-6 & 5e-6 & 5e-6 \\

Batch Size & 16 & 16 & 16 \\

Group Size & 16 & 16 & 16 \\

Training Steps & 400 & 300 & 200 \\

CE Weight & 1 & 1 & 1 \\

MSE Weight & 2 & 2 & 2\\

Temperature & 0.9 & 0.9 & 0.9 \\

Number Timesteps for Reasoning & 10 & 15 & 15  \\

Number Timesteps for Edit & 20 & 20 & 20 \\

CFG for Text & 4 & 4 & 4 \\

CFG for Image & 1.5 & 1.5 & 1.5 \\

KL for Image & 0.005 & 0.005 & 0.005 \\

KL for Text & 0.0005 & 0.0005 & 0.0005 \\
\bottomrule

\end{tabular}
\end{table}

\paragraph{GenEval++.} GenEval++~\citep{echo4o} is a more challenging extension of GenEval, featuring more complex instructions and employing advanced Vision-Language Models (VLMs) for a more robust evaluation. Following a similar procedure to GenEval, our training prompts were generated by randomly combining elements from the official templates, ensuring no overlap with the test set. For this benchmark, we employ Qwen-2.5-VL-72B as the reward model.

\paragraph{TIIF.} The TIIF benchmark~\citep{tiif} is designed to systematically assess a model's ability to interpret and follow intricate, fine-grained textual instructions. It comprises 5,000 prompts categorized into three difficulty levels, enabling a nuanced evaluation of critical capabilities such as attribute synthesis, reasoning, and style control. For our experiments, we utilize the official training set prompts for reinforcement learning and report performance on the short-prompt version of the test-mini evaluation set. The reward signal is calculated based on a list of yes/no questions with ground-truth answers provided by the benchmark. The reward $r$ is defined as the ratio of correct answers:
$$
r = \frac{N_{\text{correct}}}{N_{\text{total}}}
$$
Given model generated image, we utilize Qwen-2.5-VL-72B as the VLM to answer these questions and compute the reward.

\subsubsection{Understanding Benchmarks}
\label{appendix:understand_benchmark}
A central hypothesis of our work is that structuring the generation process through iterative reasoning with reinforcement learning not only improves the final output but also enhances the model's fundamental multimodal understanding capabilities. To empirically validate this claim, we designed two novel downstream evaluation tasks to measure the change in the Bagel model's comprehension abilities before and after training with our Reason-Reflect-Refine framework. For both tasks, we use a powerful proprietary VLM, Gemini 2.5 Flash, to establish a high-quality ground truth for comparison.

\paragraph{Compositional Visual Question Answering (VQA).} This task is designed to directly probe the model's ability to accurately perceive and verify the complex compositional elements it was trained to generate, such as object counts, attributes, and spatial relationships.
\begin{itemize}[leftmargin=*]
    \item \textbf{Data Construction.} We first utilize the Bagel model to generate a set of images based on the test prompts from the GenEval++ benchmark. Subsequently, we leverage the structured metadata of these prompts to automatically construct a corresponding set of factual, closed-ended (yes/no) questions. For instance, for a prompt describing "five cats," we generate the question, "Are there exactly 5 cats in the image?".
    \item \textbf{Evaluation Protocol.} Ground truth is established by querying Gemini 2.5 Flash with these image-question pairs and recording its answers. We conduct a careful human alignment check to ensure a high-quality annotation. We then evaluate the accuracy of both the baseline (pre-trained) Bagel model and our final (RL-trained) model on this VQA set. The primary metric is the accuracy of each model's answers when compared against the Gemini 2.5 Flash-generated ground truth. We conduct a careful human alignment check to ensure a high-quality annotation.
\end{itemize}

\paragraph{Image-Text-Alignment (ITA)} This task assesses a more holistic and nuanced form of comprehension by evaluating the model's ability to function as a reliable "judge" or reward model, a critical capability for advanced AI systems.

\begin{itemize}[leftmargin=*]
\item \textbf{Task Setup.} Using the same set of images generated for the VQA task, we prompt the Bagel model to perform an evaluation. Specifically, we employ the official GenEval++ evaluation template, which instructs the model to assess the quality and prompt-alignment of a given image.

\item \textbf{Evaluation Protocol.} We first obtain a ground-truth set of evaluations by having Gemini 2.5 Flash perform the same assessment on all image-prompt pairs. We conduct a careful human alignment check to ensure a high-quality annotation. We then compare the judgments made by the baseline and the final trained Bagel models against this ground truth. The primary metric is the agreement rate between Bagel's judgments (e.g., its scores or categorical decisions) and the assessments provided by Gemini 2.5 Flash. This allows us to quantify the improvement in the model's ability to discern high-quality, prompt-adherent generations. 
\end{itemize}


\subsubsection{Implementation Details}
We detail the specific training procedure of the R3 framework. In the Reasoning stage, we utilize a prompt batch size of 16 and sample 16 responses for each prompt. The reasoning text is sampled with temperature of 0.9, while the diffusion SDE sampling uses a noise parameter of $a=0.7$. To transition to the Reflect-Refine stage, we select a subset of 16 samples from the total pool of 256 candidates (16 prompts $\times$ 16 responses). Crucially, we employ a sampling strategy to ensure that approximately 20\% of these selected instances are "perfect" samples (i.e., achieving a reward of 1). For the subsequent Reflect-Refine stages, we maintain a text sampling temperature of 0.9 but adjust the editing diffusion noise parameter to $a=1.0$. The training trajectory length is set 2 with 1 round for reasoning and another for reflect-refine. Additional training hyperparameters are summarized in Table \ref{tab:hyperparams}.

\begin{table}[!t]
     \centering
     \caption{\textbf{Text-to-image generation ability on the GenEval benchmark~\citep{geneval}.} The best scores are \textbf{bolded}.}
     \resizebox{0.99\linewidth}{!}{
         \begin{tabular}{lcccccccc}
             \toprule
             Method & \multicolumn{1}{c}{Single object} & \multicolumn{1}{c}{Two object} & \multicolumn{1}{c}{Counting} & \multicolumn{1}{c}{Colors} & \multicolumn{1}{c}{Position} & \multicolumn{1}{c}{Color attribution} & \multicolumn{1}{c}{Overall} \\
             \midrule
             FLUX.1-dev~\citep{FLUX} & 0.99 & 0.81 & 0.79 & 0.74 & 0.20 & 0.47 & 0.67 \\
             \midrule
             Qwen-Image-RL~\citep{qwenimage} & 1.00 & 0.95 & 0.93 & 0.92 & 0.87 & 0.83 & 0.91 \\
             FlowGRPO~\citep{flowgrpo} & 1.00 & 0.99 & 0.95 & 0.92 &  \textbf{0.99} &  0.86 & 0.95 \\
             \midrule
             BAGEL~\citep{bagel} & 0.99 & 0.94 & 0.81 & 0.88 & 0.64 & 0.63 & 0.82 \\
             BAGEL + Ours  & \textbf{1.00} & \textbf{1.00} & \textbf{0.95} & \textbf{0.95} & 0.98 & \textbf{0.89} & \textbf{0.96} \\
             \bottomrule
         \end{tabular}
     }
     \label{tab:geneval}
\end{table}

\subsection{Effect of RL training}

To clarify the effect of RL training, we conduct an ablation study comparing Bagel (without RL training) against our RL-trained model using the R3 framework. The results are presented in \cref{tab:inference_comparison}.

\begin{table}[h]
\centering
\small
\begin{tabular}{lcc}
\toprule
\textbf{Inference Strategy} & \textbf{Bagel} & \textbf{Ours} \\
\midrule
Reasoning only & 0.399 & 0.593 \\
Reason+RR $\times$ 1 & 0.432 & 0.675 \\
Reason+RR $\times$ 2 & 0.436 & 0.689 \\
Reason+RR $\times$ 3 & 0.439 & 0.682 \\
Reason+RR $\times$ 4 & 0.439 & 0.689 \\
Reason+RR $\times$ 5 & 0.439 & - \\
\bottomrule
\end{tabular}
\caption{Inference performance on GenEval++ comparison between Bagel and R3 (ours) under same inference strategies.}
\label{tab:inference_comparison}
\end{table}

These results demonstrate the effectiveness of RL training:
\begin{enumerate}[leftmargin=*]
    \item Achieve higher performance ceiling: Our RL-trained model achieves substantially better performance across all iteration counts (0.593 vs. 0.399 with reasoning only, and 0.689 vs. 0.439 at convergence), indicating that RL fundamentally enhances the model's capability to understand and generate improved responses.
     \item Improve efficiency: Our model converges to near-optimal performance within 2 Reflection-Refine rounds (0.689), whereas Bagel requires 3 rounds to reach its plateau (0.439). This demonstrates that RL training not only elevates the performance ceiling but also accelerates convergence, making the approach more computationally efficient in practice.
\end{enumerate}

Notably, the performance gap between our model and Bagel increases with the Reflection-Refine framework, suggesting that RL training better equips the model to leverage iterative refinement effectively.

\subsection{Computational Cost Analysis}
\label{sec:cost_analysis}
A key practical concern for iterative refinement is computational overhead. We provide a computational cost analysis as follows:

\paragraph{Adaptive Inference.} Unlike fixed-iteration approaches, R3 learns to self-terminate when generation quality is satisfactory. On GenEval++, the distribution of refinement iterations shows efficient resource allocation: 45\% of prompts finish immediately (0 refinement cost), 26\% require 1 refinement turn, 14\% need 2 refinement turns, and only 15\% require 3+ refinement turns. This adaptive behavior significantly reduces average computational cost.

\paragraph{Inference Latency.} We measure wall-clock time on a single NVIDIA H20 GPU (batch size 1, 512$\times$512 resolution). The initial Reasoning stage takes 20--25s. Each Reflect-Refine turn requires 25--35s, with Reflection (text) accounting for 5--10s and Refinement (image) for 20--25s.

\subsection{Visualization on multi-round generation}
\label{app:mr_vis}
\begin{figure}
    \centering
    \includegraphics[width=0.99\linewidth]{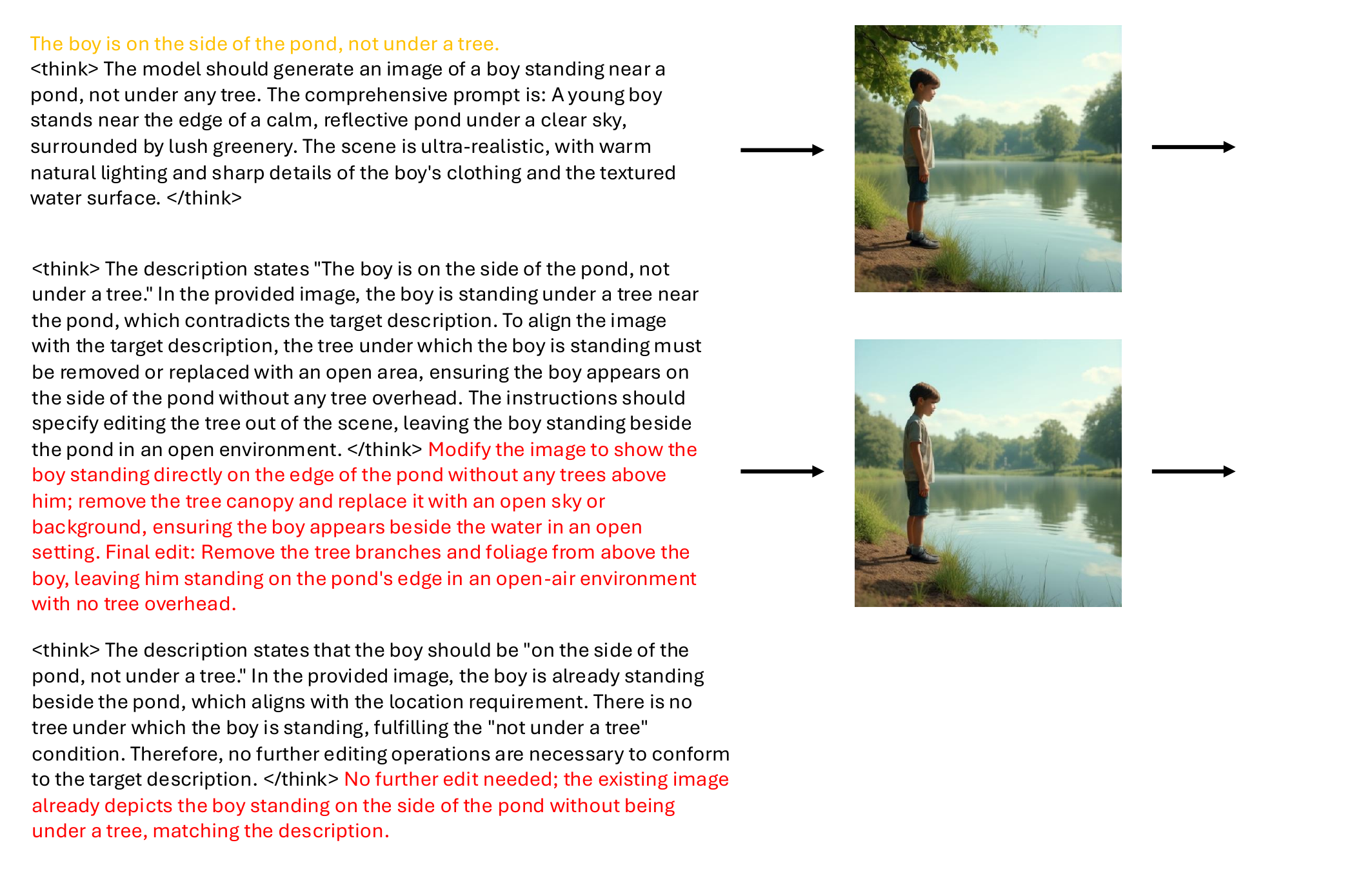}
    \caption{An illustration of our model's two-stage generation process. For the prompt "The boy is on the side of the pond, not under a tree," the model's initial reasoning stage erroneously generates a tree above the boy. The subsequent reflect-refine stage corrects this error, after which the process terminates automatically.}
    \label{fig:mr_boy}
\end{figure}

\begin{figure}
    \centering
    \includegraphics[width=0.99\linewidth]{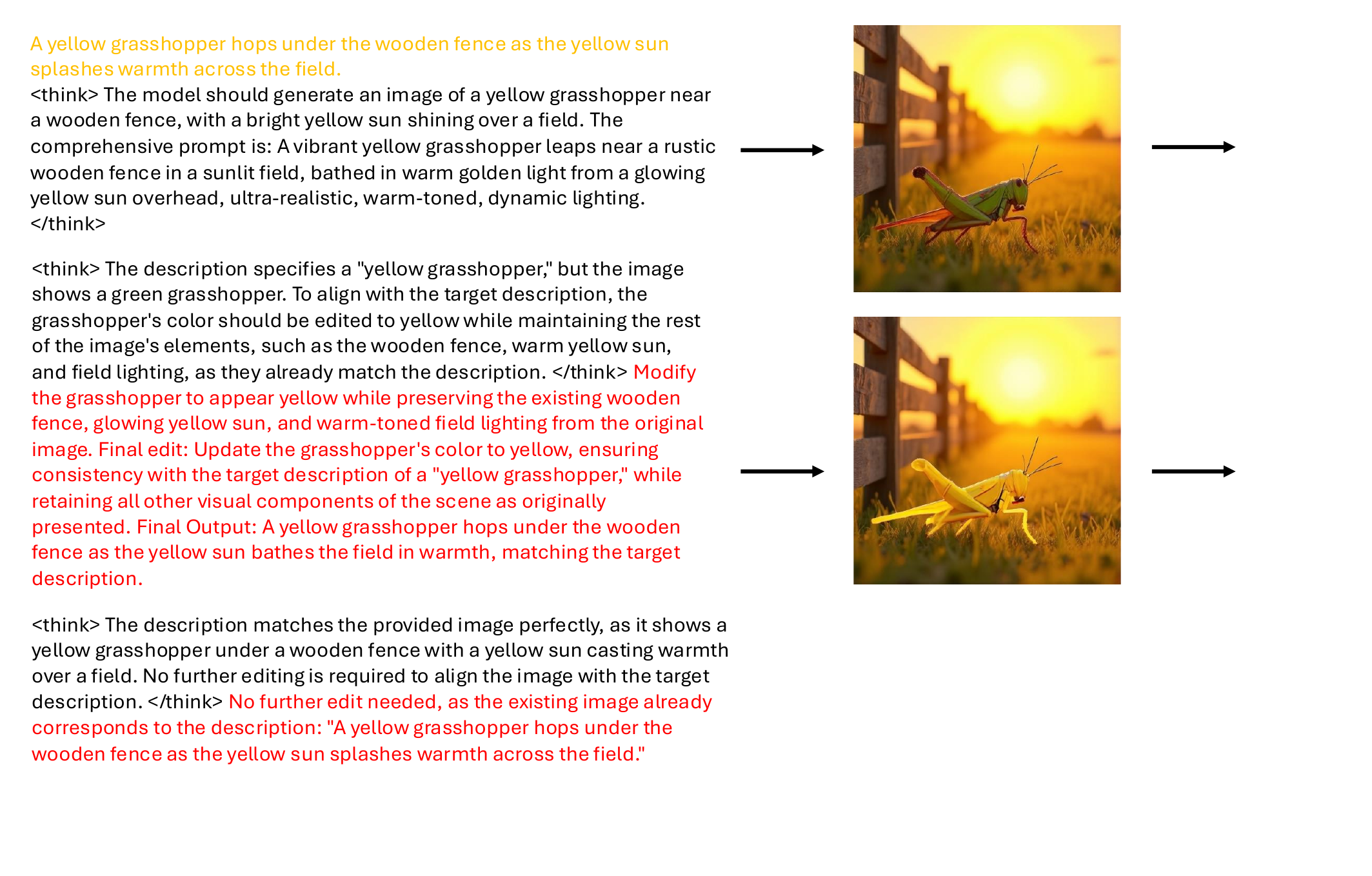}
    \caption{An illustration of our model's two-stage generation process. For the prompt "A yellow grasshopper hops under the wooden fence as the yellow sun splashes warmth across the field," the model's initial reasoning stage generates a grasshopper with an incorrect color. The subsequent reflect-refine stage corrects the color, after which the process terminates automatically.}
    \label{fig:mr_grasshopper}
\end{figure}

\begin{figure}
    \centering
    \includegraphics[width=0.99\linewidth]{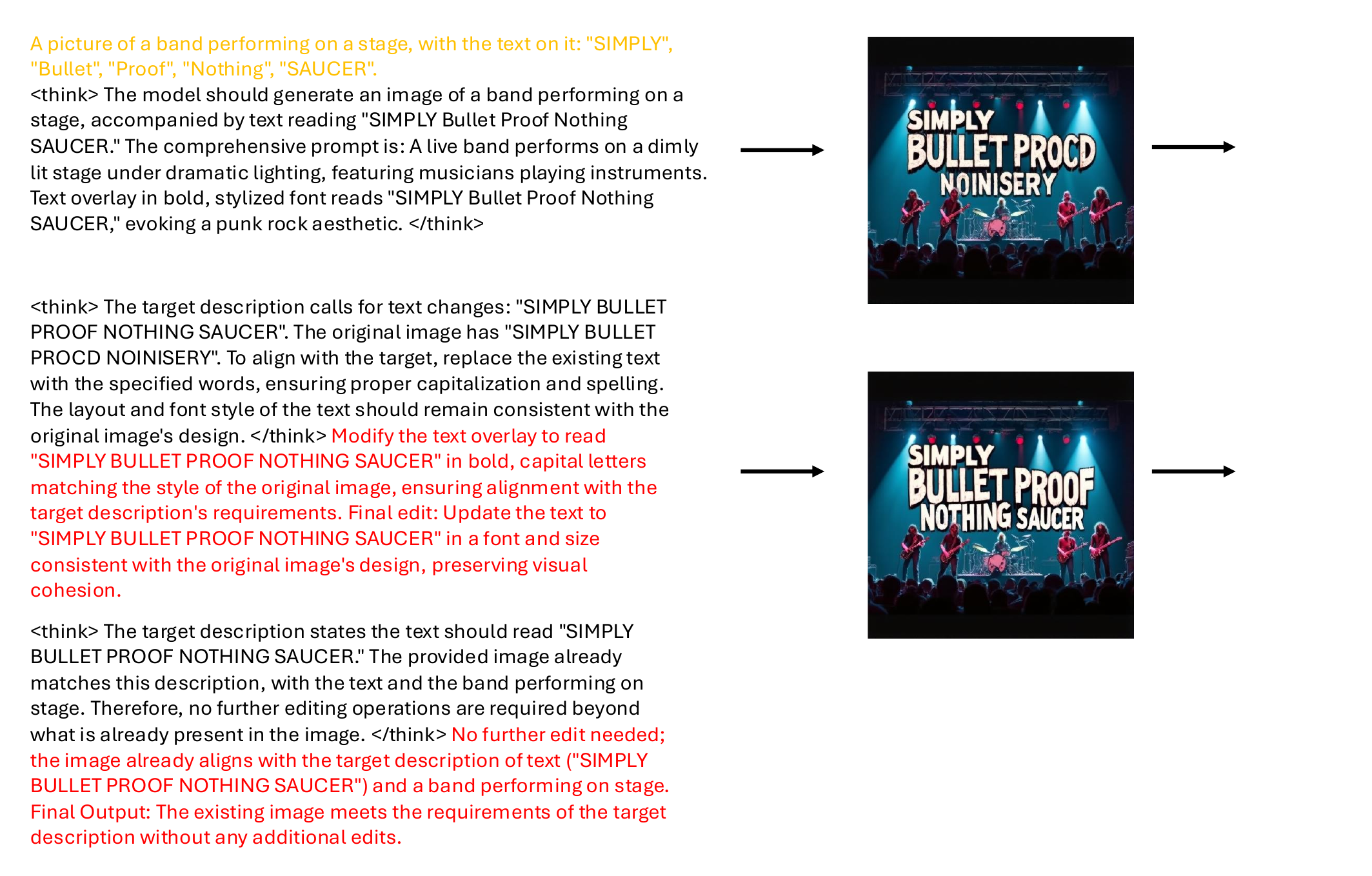}
    \caption{An illustration of our model's two-stage generation process. For the prompt 'A picture of a band performing on a stage, with the text on it: "SIMPLY", "Bullet", "Proof", "Nothing", "SAUCER"', the model's initial stage fails to correctly render some of the text. The subsequent reflect-refine stage successfully corrects the text to match the prompt, after which the process terminates automatically.}
    \label{fig:mr_text2}
\end{figure}

\begin{figure}
    \centering
    \includegraphics[width=0.99\linewidth]{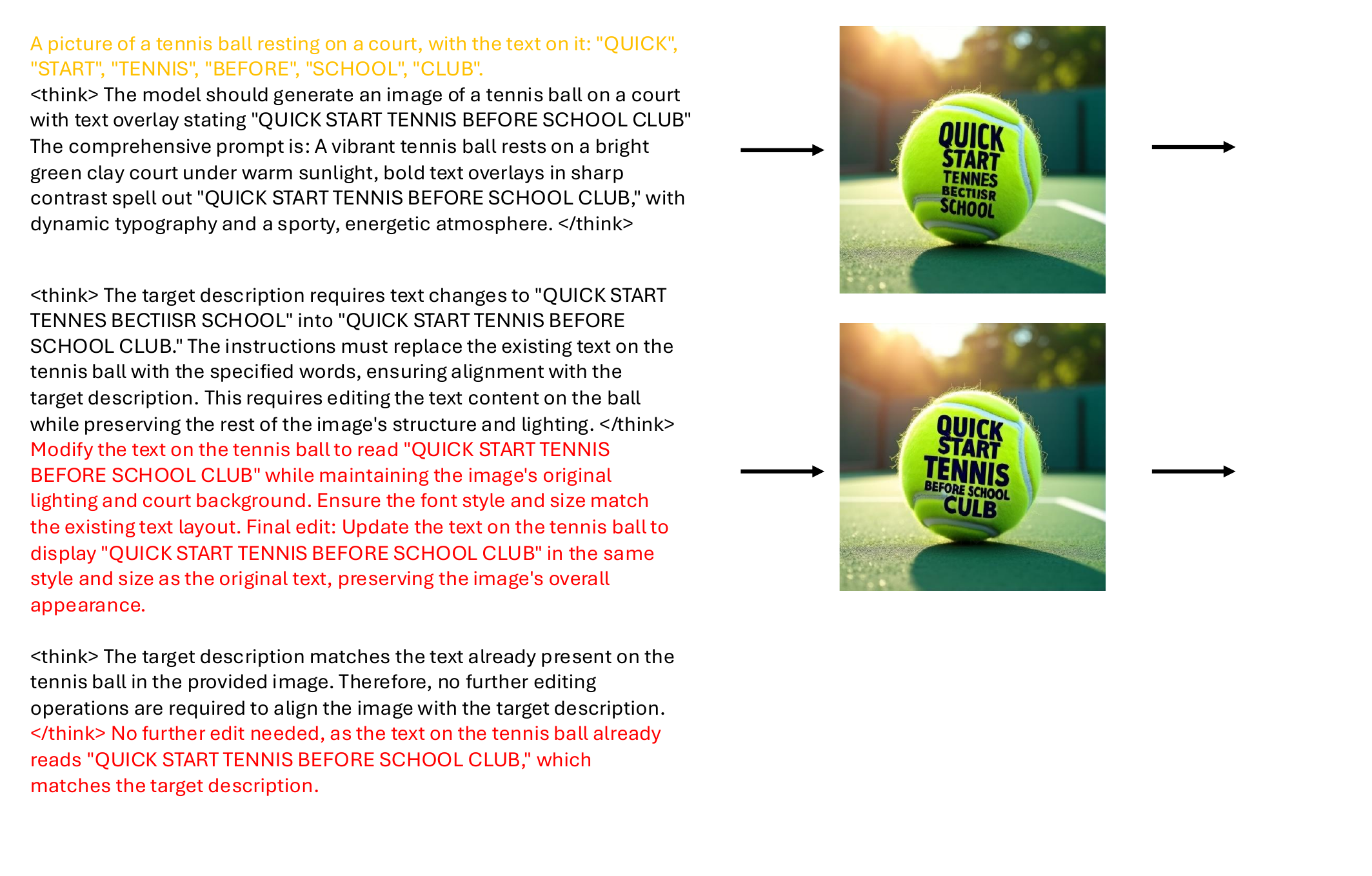}
    \caption{An illustration of our model's two-stage generation process, showing a limitation. For the prompt requiring specific text on a tennis ball, the initial stage fails to render all the words. The reflect-refine stage adds most of the missing text but introduces a minor spelling error ("CLUB"). Despite this inaccuracy, the model terminates the process without attempting further corrections.}
    \label{fig:mr_baseball_text}
\end{figure}

\begin{figure}
    \centering
    \includegraphics[width=0.99\linewidth]{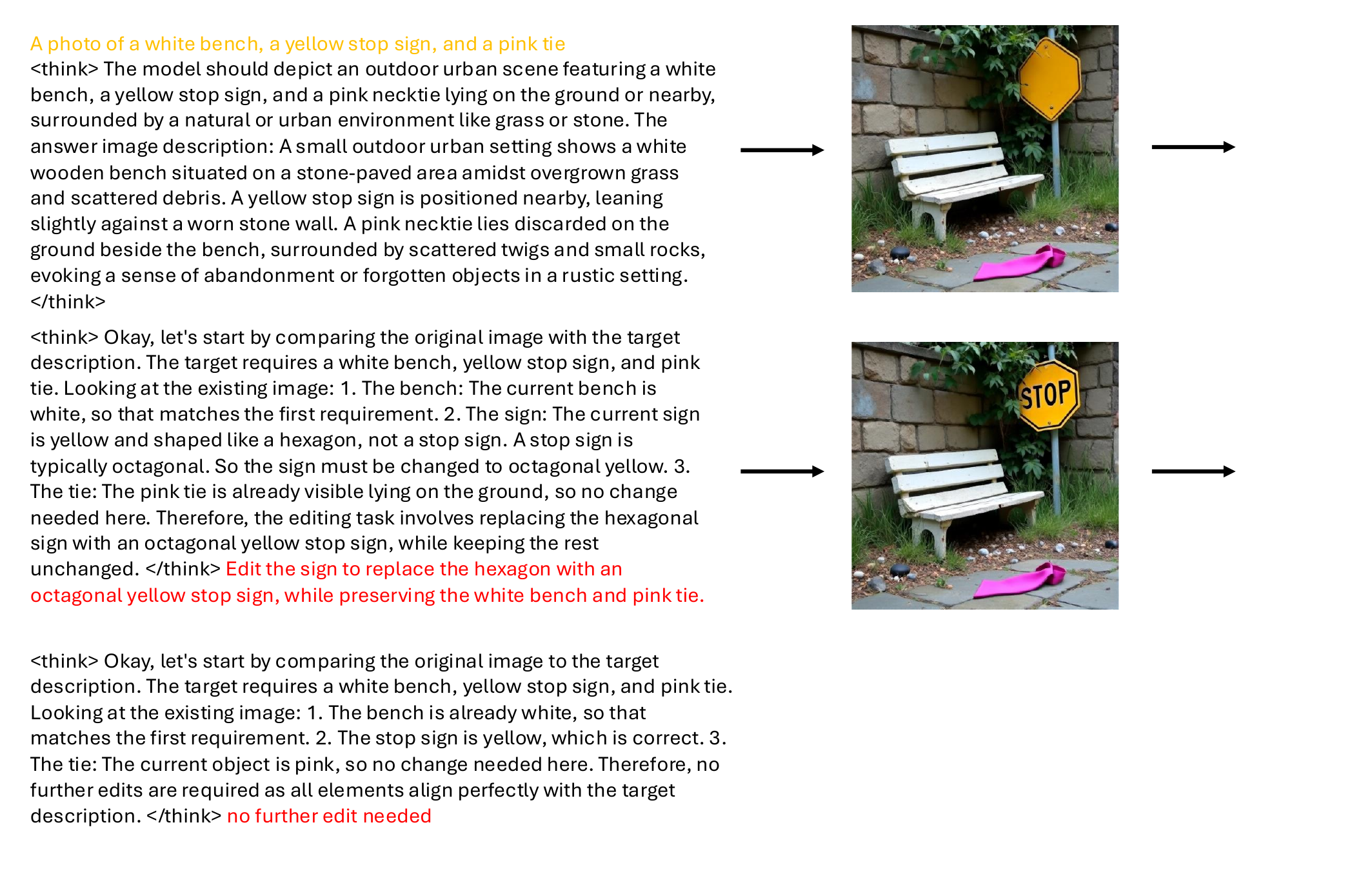}
    \caption{An illustration of our model's two-stage generation process. For the prompt "A photo of a white bench, a yellow stop sign, and a pink tie," the model's initial generation fails to correctly render the stop sign. The subsequent reflect-refine stage successfully generates the object as specified, and the process terminates automatically.}
    \label{fig:mr_stop_sign}
\end{figure}

\begin{figure}
    \centering
    \includegraphics[width=0.99\linewidth]{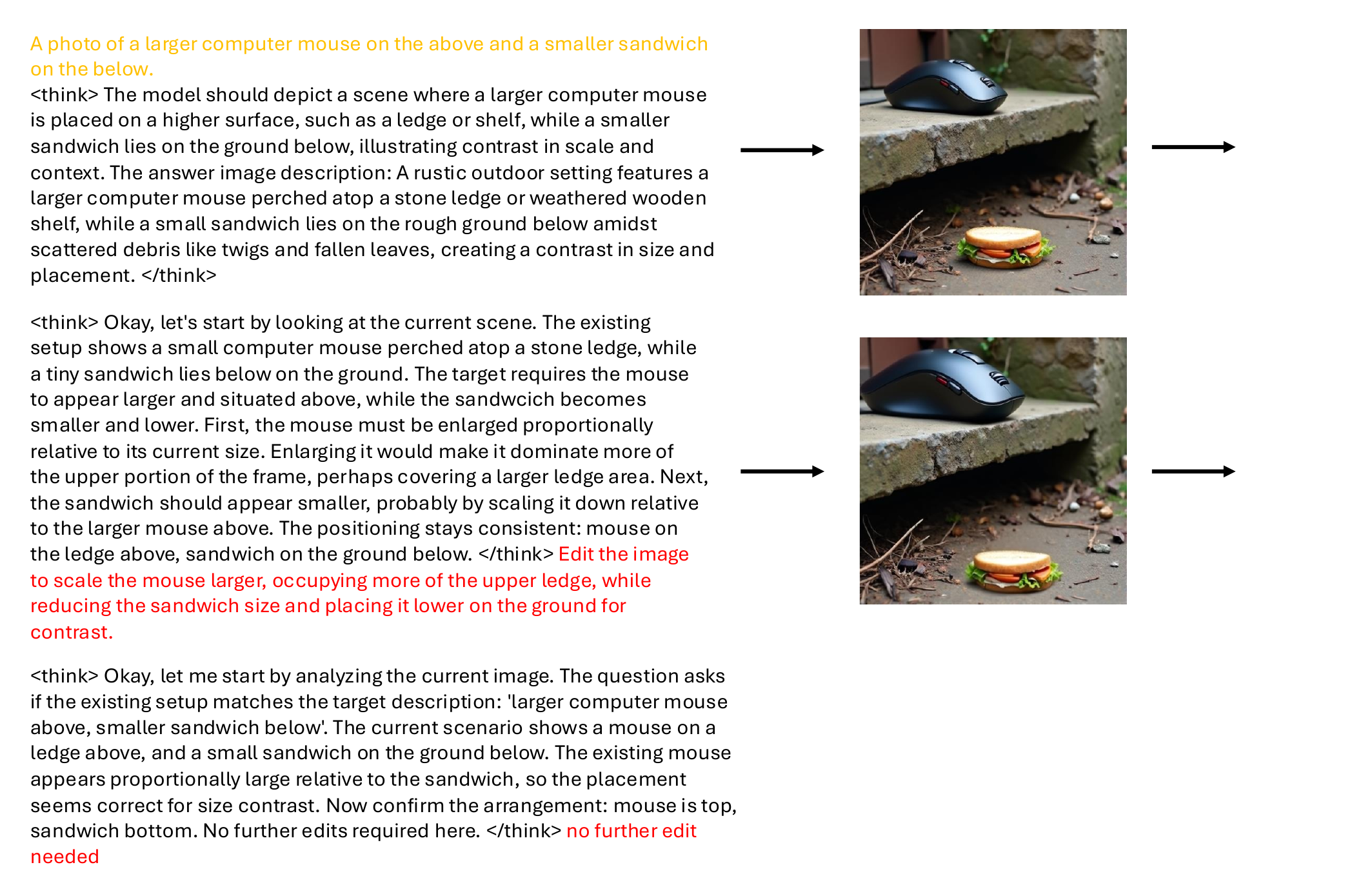}
    \caption{An illustration of our model's two-stage generation process. For the prompt "A photo of a larger computer mouse on the above and a smaller sandwich," the model's initial generation fails to correctly depict the specified size relationship. The subsequent reflect-refine stage adjusts the relative sizes of the objects to match the prompt, after which the process terminates.}
    \label{fig:mr_mouse}
\end{figure}

\begin{figure}
    \centering
    \includegraphics[width=0.99\linewidth]{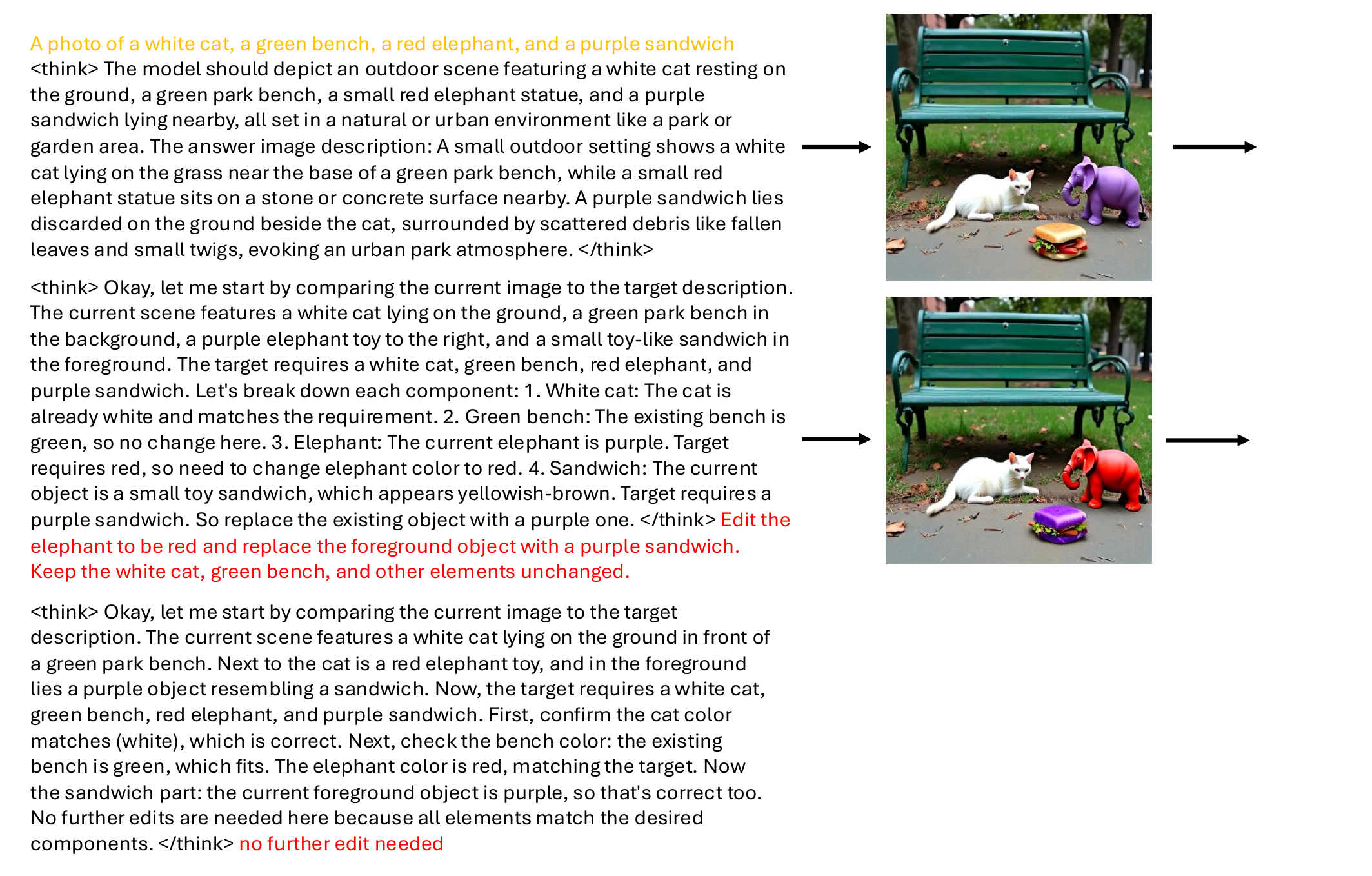}
    \caption{An illustration of our model's two-stage generation process. For the prompt "a photo of a white cat, a green bench, a red elephant, and a purple sandwich on the below," the model's initial generation renders both the elephant and the sandwich with incorrect colors. The subsequent reflect-refine stage corrects the colors of both objects, and the process terminates automatically.}
    \label{fig:mr_color}
\end{figure}

\begin{figure}
    \centering
    \includegraphics[width=\linewidth]{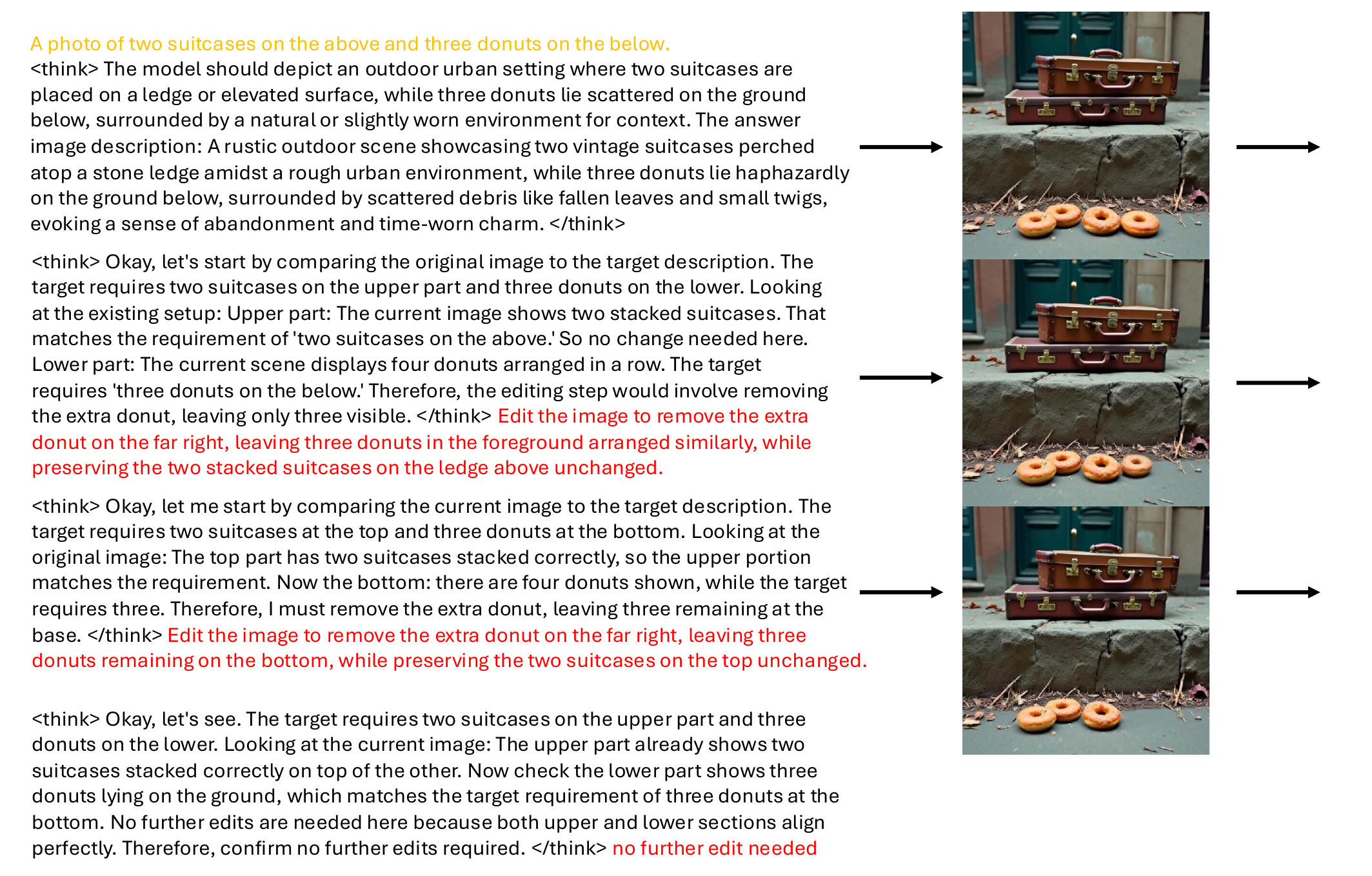}
    \caption{An illustration of our model's three-stage generation process. For the prompt "A photo of two suitcases on the above and three donuts on the below," the initial generation incorrectly produces four donuts. In the first reflect-refine stage, the model identifies the count error but fails to correct it. A second reflect-refine stage is initiated, where the model successfully edits the image to show the correct number of donuts, after which the process terminates automatically.}
    \label{fig:mr_donut}
\end{figure}

We demonstrate the multi-round editing process of our proposed R3 framework in~\cref{fig:mr_boy,fig:mr_grasshopper,fig:mr_text2,fig:mr_baseball_text,fig:mr_stop_sign,fig:mr_mouse,fig:mr_color,fig:mr_donut}. The figure details each stage, including the initial Reason stage's planning text, the subsequent reflection text, and the resulting refined image.

\newpage
\subsection{Extension of R3 framework to Maze Navigation Task}
To demonstrate the versatility of our proposed R3 (Reason-Reflect-Refine) framework beyond text-to-image generation, we apply it to the complex task of Maze Navigation. In this task, the model is given an initial image of a maze and is required to plot a solution path from the entrance to the exit. Generating the entire path in a single forward pass is challenging for complex mazes; therefore, the iterative nature of our R3 framework is particularly well-suited for decomposing this problem into a sequence of manageable sub-steps.

Our training strategy consists of two stages. First, we perform Supervised Fine-Tuning (SFT) on a dataset of random trajectories within various mazes. This initial stage equips the model with the fundamental capability of state transition—that is, generating the subsequent maze image based on its current state and a given movement instruction. Subsequently, we employ Reinforcement Learning (RL) to optimize the framework, enabling the model to learn an effective policy for generating a sequence of movement instructions that successfully navigates the maze from start to finish.
\begin{figure}
    \centering
    \includegraphics[width=\linewidth]{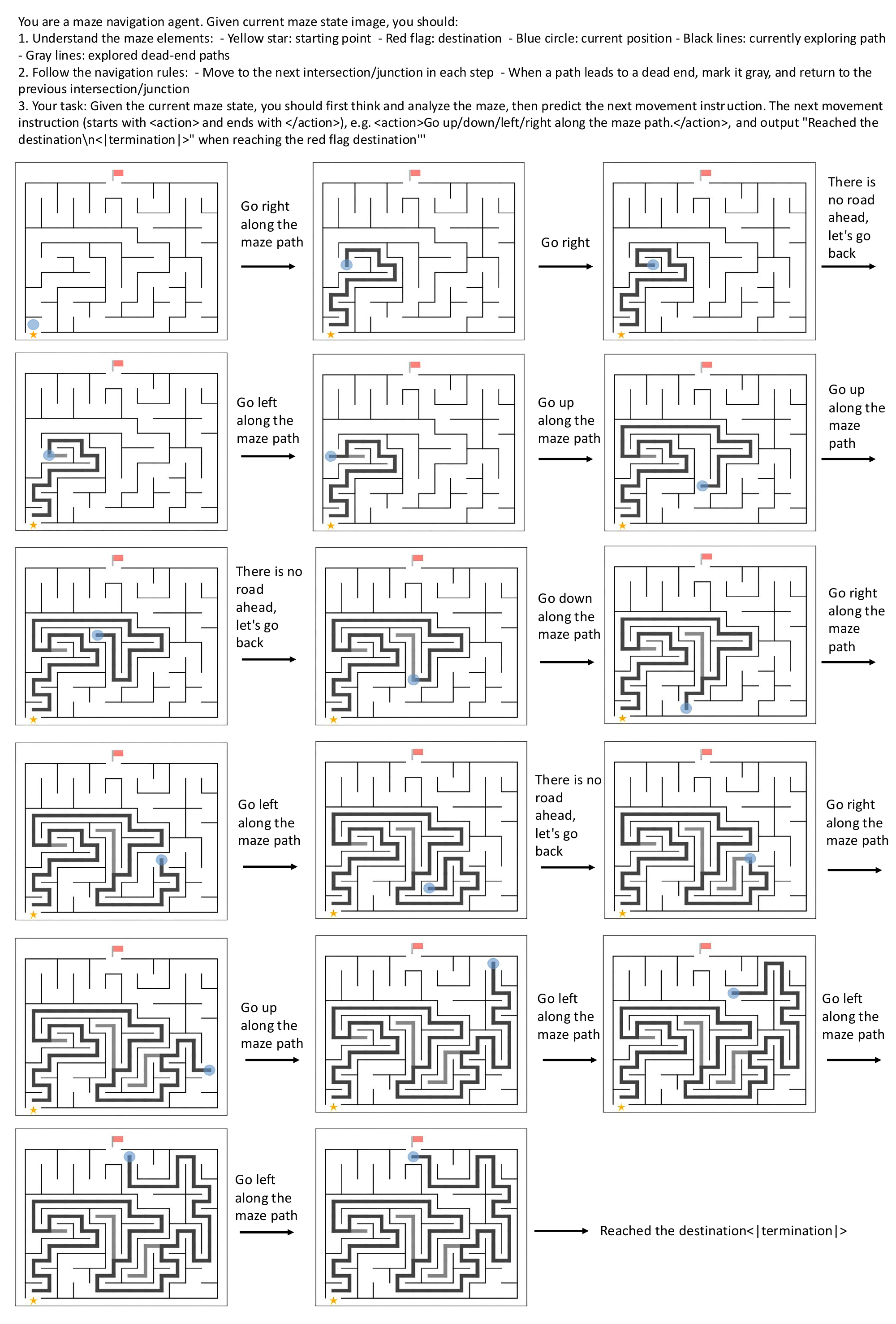}
    \caption{An illustration of maze navigation results. The model is tested on a 12x10 maze, where it demonstrates the ability to accurately identify valid directions during the navigation process. When faced with multiple paths or dead ends, the model can effectively backtrack. Furthermore, it automatically recognizes when the destination has been reached and terminates the process accordingly.}
    \label{fig:placeholder}
\end{figure}
\end{document}